\documentclass[lettersize,journal]{IEEEtran}
\usepackage{amsmath,amsfonts}
\usepackage{algorithm}
\usepackage{array}
\usepackage[caption=false,font=normalsize,labelfont=sf,textfont=sf]{subfig}
\usepackage{textcomp}
\usepackage{stfloats}
\usepackage{url}
\usepackage{verbatim}
\usepackage{graphicx}
\usepackage{cite}

\usepackage{tcolorbox}
\usepackage{booktabs}
\usepackage{listings}
\usepackage{xcolor}
\usepackage{enumitem}
\tcbuselibrary{listings,skins,breakable}

\definecolor{codegreen}{rgb}{0,0.6,0}
\definecolor{codegray}{rgb}{0.5,0.5,0.5}
\definecolor{codepurple}{rgb}{0.58,0,0.82}
\definecolor{backcolour}{rgb}{0.95,0.95,0.92}

\usepackage{booktabs}
\usepackage{multirow}
\usepackage{breqn}
\usepackage{algpseudocode}

\algnewcommand\Input{\item[\textbf{Input:}]}
\algnewcommand\Output{\item[\textbf{Output:}]}

\newcommand{\eat}[1]{}

\usepackage{xcolor}

\begin{document}

\title{LLM-ODDR: A Large Language Model Framework for Joint Order Dispatching and Driver Repositioning}

\author{
Tengfei Lyu, Siyuan Feng, Hao Liu, and Hai Yang

\thanks{Tengfei Lyu is with the Thrust of Artificial Intelligence, The
Hong Kong University of Science and Technology (Guangzhou), Guangzhou, China, and The Hong Kong University of Science and Technology, Hong Kong, China
(e-mail: tlyu077@connect.hkust-gz.edu.cn).}

\thanks{Siyuan Feng is with the Department of Aeronautical and Aviation Engineering, The Hong Kong Polytechnic University, Hong Kong, China and Research Center for Low Altitude Economy, The Hong Kong Polytechnic University, Hong Kong, China
(e-mail: siyuan.feng@polyu.edu.hk).}

\thanks{Hao Liu is with the Thrust of Artificial Intelligence, The
Hong Kong University of Science and Technology (Guangzhou), Guangzhou, China, and also with the Department of Computer Science and Engineering,  The Hong Kong University of Science and Technology, Hong Kong, China 
(e-mail: liuh@ust.hk).}

\thanks{Hai Yang is with the Department of Civil and Environmental Engineering, The
Hong Kong University of Science and Technology, Hong Kong, China
(e-mail: cehyang@ust.hk).}

\thanks{Corresponding authors: Siyuan Feng and Hao Liu.}
}

\maketitle

\begin{abstract}
Ride-hailing platforms face significant challenges in optimizing order dispatching and driver repositioning operations in dynamic urban environments. Traditional approaches based on combinatorial optimization, rule-based heuristics, and reinforcement learning often overlook driver income fairness, interpretability, and adaptability to real-world dynamics. To address these gaps, we propose LLM-ODDR, a novel framework leveraging Large Language Models (LLMs) for joint Order Dispatching and Driver Repositioning (ODDR) in ride-hailing services. LLM-ODDR framework comprises three key components: (1) Multi-objective-guided Order Value Refinement, which evaluates orders by considering multiple objectives to determine their overall value; (2) Fairness-aware Order Dispatching, which balances platform revenue with driver income fairness; and (3) Spatiotemporal Demand-Aware Driver Repositioning, which optimizes idle vehicle placement based on historical patterns and projected supply. We also develop JointDR-GPT, a fine-tuned model optimized for ODDR tasks with domain knowledge. Extensive experiments on real-world datasets from Manhattan taxi operations demonstrate that our framework significantly outperforms traditional methods in terms of effectiveness, adaptability to anomalous conditions, and decision interpretability. To our knowledge, this is the first exploration of LLMs as decision-making agents in ride-hailing ODDR tasks, establishing foundational insights for integrating advanced language models within intelligent transportation systems. While the current framework incurs higher computational costs than traditional methods, we show that parallel decomposition and model distillation can reduce latency to production-viable levels for deployment.
\end{abstract}


\section{Introduction}
\label{Introduction}

\begin{figure}[t]
  \centering
  \includegraphics[width=1\linewidth]{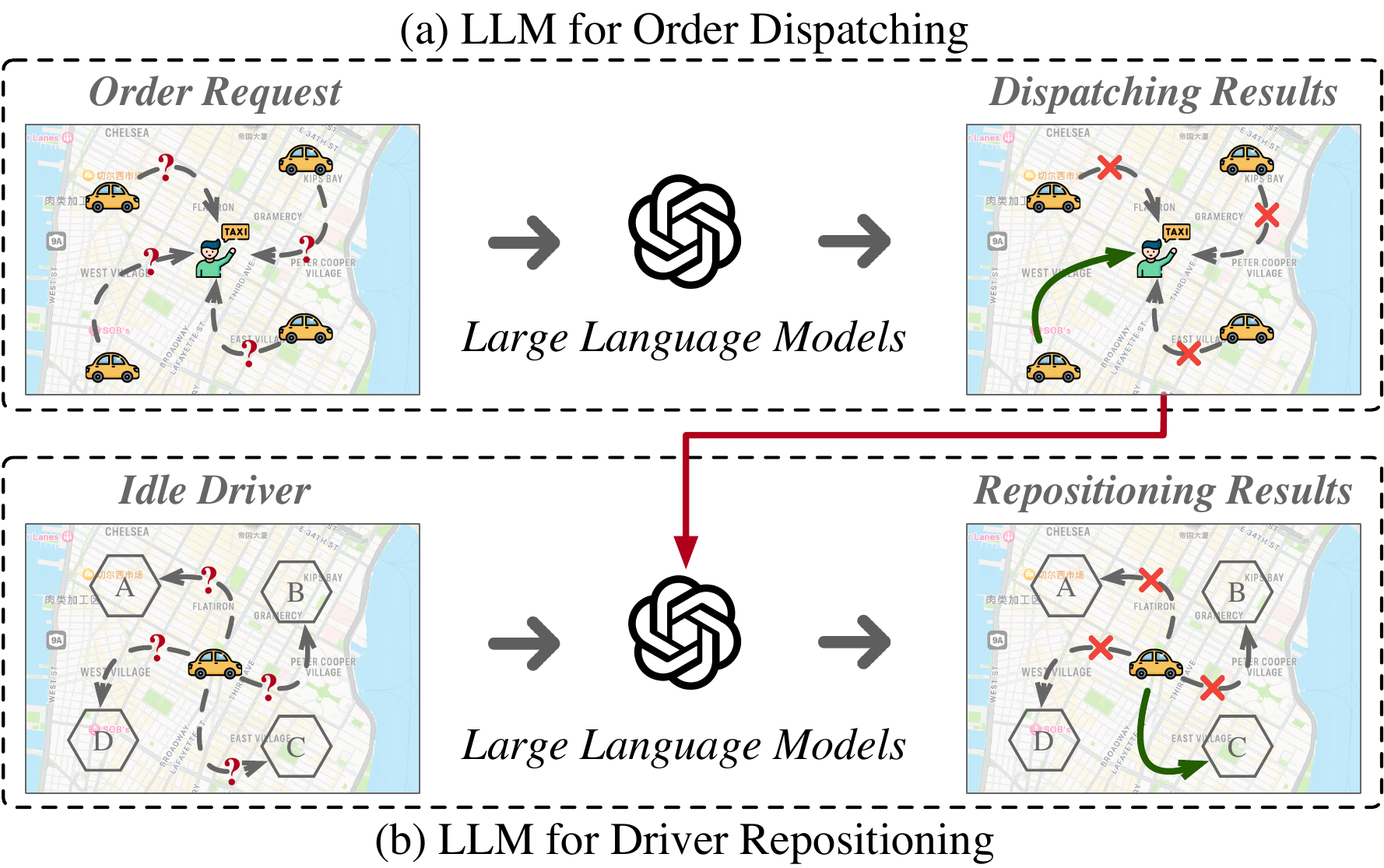}
  \caption{The pipeline of large language models for \textbf{joint order dispatching and driver repositioning} tasks.}
  \label{fig:introduction}
\end{figure}

Ride-hailing platforms \cite{tirachini2020ride,suhr2019two} have revolutionized urban transportation by efficiently connecting passengers with vehicles through digital marketplaces. These platforms face complex real-time decision-making challenges, particularly in order dispatching (matching riders to drivers) and driver repositioning (strategically relocating idle vehicles) \cite{liu2022machine}. The effectiveness of these operations fundamentally determines platform sustainability, customer satisfaction, fleet efficiency, and driver retention. Traditional approaches based on combinatorial optimization \cite{tong2021combinatorial}, rule-based heuristics \cite{ma2013t}, and reinforcement learning (RL) \cite{jin2019coride,yu2021deep,wei2023reinforcement,yue2023reinforcement,liu2025learning,hu2025bmg} often struggle with the multi-dimensional complexity and dynamic nature of real-world environments.

Traditional optimization methods, despite providing theoretical guarantees, often fail to handle the dynamic and unpredictable nature of urban ride-hailing scenarios \cite{liu2022machine, zhang2017taxi}, as they typically rely on rigid formulations. Rule-based heuristics lack flexibility for rapidly evolving conditions. Recent RL-based approaches \cite{xu2018large, lin2018efficient, jin2019coride, wei2023reinforcement, yue2023reinforcement} show improved responsiveness by learning adaptive policies; however, they generally focus on single-objective optimization, limiting their practical effectiveness when balancing multiple competing goals.
Furthermore, there is growing recognition that addressing order dispatching and driver repositioning (ODDR) as separate problems is suboptimal. The joint optimization of ODDR represents a more practical approach \cite{jin2019coride, sun2024optimizing, sun2022optimizing}, acknowledging the inherent interdependence between these decisions: current dispatch choices affect future vehicle distributions, while repositioning strategies directly impact dispatch possibilities.
Despite these advancements, significant challenges persist in current RL-based ODDR methods. These approaches typically optimize for singular objectives (e.g., maximizing revenue or minimizing wait times), overlooking the inherent trade-offs and complex interdependencies between these goals. For instance, aggressively maximizing immediate platform revenue might sacrifice driver utilization efficiency, leading to driver dissatisfaction and eventual supply shortage. Similarly, prioritizing passenger wait time reduction often requires maintaining excess vehicle supply in high-demand areas, potentially reducing overall system efficiency and driver earnings. Furthermore, these methods struggle to incorporate multiple constraints simultaneously, lack interpretability in their decision-making processes, and demonstrate limited adaptability to abnormal conditions such as surge events. The absence of a holistic approach that recognizes these interrelationships significantly hinders the development of sustainable ride-hailing systems capable of balancing competing stakeholder interests while maintaining operational resilience under varying conditions.

Large Language Models (LLMs) have recently emerged as powerful tools for reasoning \cite{mondorfbeyond,wang2024rethinking}, decision-making \cite{lai2024llmlight,yu2024fincon,jiawei2024large}, and processing complex multimodal information \cite{li2024urbangpt,le2024multimodal}. Their natural language understanding capabilities, contextual reasoning, and ability to synthesize diverse inputs position them as promising solutions for addressing multifaceted challenges \cite{zhang2024large}. 
While initial applications of LLMs in transportation primarily focused on natural language interfaces \cite{meng-etal-2024-llm, song2023llm}, recent advancements have catalyzed their evolution into core decision-making agents \cite{lai2024llmlight, han2024gpt}. 
For instance, LLMLight \cite{lai2024llmlight} employs a knowledge-enhanced prompt that details real-time traffic conditions, enabling LLMs to function as decision-making agents for traffic signal control. 
In addition, TransCompressor \cite{yang2024transcompressor} is a novel framework that leverages Large Language Models to efficiently compress and decompress multimodal transportation sensor data.
In ride-hailing services, GARLIC \cite{GARLIC2025aaai} leverages a transformer network to augment reinforcement learning and intelligent control for vehicle dispatching, enhancing prediction accuracy and optimizing dispatching policies in real-world scenarios.
The contextual reasoning ability of LLMs and their capacity to synthesize diverse inputs position them as particularly promising solutions for simultaneously addressing the multiple competing objectives and constraints that characterize ride-hailing operations \cite{zhang2024large}.

The integration of LLMs into ride-hailing services represents a paradigm shift in complex transportation decision-making. Unlike traditional methods that operate within rigid optimization frameworks with predefined objectives, LLMs offer several distinct advantages for ODDR tasks.
LLMs excel at multi-dimensional reasoning, inherently balancing competing operational priorities through their sophisticated analytical capabilities. This enables simultaneous consideration of immediate revenue maximization alongside long-term strategic factors such as future area value, vehicle utilization efficiency, and driver income fairness, a holistic optimization approach that conventional methods struggle to achieve without complex formulation.
Traditional methods \cite{jin2019coride,yang2024rethinking,xu2018large,lin2018efficient} typically require extensive parameter tuning for different operational scenarios (e.g., rush hours vs. late nights, weekdays vs. weekends). In contrast, LLMs demonstrate remarkable contextual adaptability, leveraging their pre-trained knowledge to generalize effectively across diverse scenarios. This is particularly valuable for anomalous events like concert dismissals or sporting events, where LLMs can reason about expected demand patterns despite limited historical data for such specific occurrences.
Moreover, LLMs also transform the historically opaque "black-box" nature of algorithmic decision-making in transportation systems into transparent, explainable processes. This transparency builds stakeholder trust and enables targeted system refinements based on explicit reasoning chains rather than abstract performance metrics alone.
Recent studies \cite{lai2024llmlight,zhang2024large,liu2024spatial,movahedi2024crossroads,zheng2023trafficsafetygpt,liu2024large} highlight LLMs' evolution from mere interfaces to decision-making agents in transportation, though significant challenges remain in their application to ride-hailing operations.

Despite their promising capabilities, integrating LLMs into ride-hailing operations presents two significant challenges.
First, the fundamental challenge in joint ODDR tasks lies in understanding and modeling the complex interdependencies between dispatching and repositioning decisions. Unlike separate optimization approaches, LLMs must capture how current dispatch choices create future vehicle distributions that constrain repositioning options, while repositioning strategies directly reshape the potential dispatch landscape. This causal entanglement requires sophisticated spatio-temporal reasoning capabilities beyond traditional sequential processing—LLMs must effectively reason about these spatio-temporal dependencies to generate coherent joint policies rather than locally optimal but globally suboptimal separate decisions.
Second, ODDR decisions demand precise quantitative reasoning across multiple competing objectives (e.g., platform revenue, wait time minimization, driver fairness). While LLMs demonstrate strong qualitative reasoning, their quantitative precision, especially when evaluating trade-offs between objectives with different units and scales, remains a significant challenge. This necessitates developing specialized prompting strategies or intermediate reasoning frameworks that guide LLMs to perform consistent numerical analyses across diverse operational scenarios.
Overcoming these challenges is essential to realize the full potential of LLMs in advancing intelligent transportation systems.

To address these challenges, we propose LLM-ODDR, a novel LLM-based framework for joint order dispatching and driver repositioning (ODDR) in ride-hailing services, as depicted in Figure \ref{fig:introduction}. Our framework combines three key components that harness the reasoning capabilities of LLMs: (1) Multi-objective-guided Order Value Refinement, which enables comprehensive order evaluation by integrating immediate rewards with future area value through an iterative self-refinement mechanism; (2) Fairness-aware Order Dispatching, which balances platform revenue with driver income fairness to ensure sustainable operations; (3) Spatiotemporal Demand-Aware Driver Repositioning, which optimizes idle vehicle placement by synthesizing historical demand patterns with projected vehicle supply. This framework overcomes the limitations of existing approaches by simultaneously addressing multiple optimization objectives while providing interpretable decision rationales.
Finally, we perform hybrid instruction fine-tuning to construct a specialized LLM JointDR-GPT based on Llama 3.1-70B by using LoRA \cite{hulora}, to enhance the LLM-ODDR framework.
We acknowledge that LLM-based decision-making currently incurs higher computational overhead than traditional optimization and reinforcement learning methods. However, as discussed in Section V-F and Appendix E, this challenge can be effectively mitigated through parallel inference decomposition, model distillation, and other deployment optimization techniques, making practical adoption a realistic near-term prospect.

In summary, this paper makes the following contributions:
\begin{itemize}
    \item we introduce the LLM-ODDR framework \footnote{\url{https://github.com/usail-hkust/LLM-ODDR}} that jointly optimizes order dispatching and driver repositioning, offering new capabilities for ride-hailing services. To our knowledge, this is the first exploration of LLMs as decision-making agents in ODDR tasks.
    
    \item We develop the fine-tuned JointDR-GPT, optimized for joint ODDR tasks, which significantly enhances the framework’s ability to generate effective decisions.  Specifically, we design specialized prompting strategies and domain knowledge to enable LLMs to reason about order values, driver fairness, and spatiotemporal demand patterns, which establish critical precedents for LLM integration in ride-hailing platform.  

    \item Extensive experiments on real-world datasets from Manhattan taxi operations demonstrate the superior efficacy and interpretability of our framework. 
    These experimental findings establish foundational insights into how LLMs can be strategically integrated within intelligent transportation systems to optimize performance.
\end{itemize}

The remainder of this paper is organized as follows: Section \ref{Related Works} reviews pertinent literature on decision-making in ride-hailing systems and LLM applications in transportation. Section \ref{Problem Formulation} formally defines the joint order dispatching and driver repositioning problem. Section \ref{Methods} presents our proposed LLM-ODDR framework in detail. Section \ref{Experiments} describes our experimental setup and presents comprehensive results, followed by discussion and conclusions in Section \ref{Conclusion}.

\section{Related Works}
\label{Related Works}
In this section, we briefly review related works most pertinent to our study, focusing on two key aspects: decision-making in ride-hailing systems, and large language models in decision-making for transportation systems.

\subsection{Decision-Making in Ride-Hailing Systems}
Prior research has addressed two core challenges in ride-hailing systems: order dispatching (matching riders to drivers) and driver repositioning (guiding idle drivers to high-demand areas). The primary objectives focus on maximizing profit \cite{wang2019adaptive,zhang2017taxi,wang2024spatial,tang2021value}, minimizing travel costs \cite{tong2016online,wang2018stable}, reducing passenger waiting times \cite{xu2019unified}, optimizing fairness \cite{feng2024multi,sun2022optimizing,sun2024optimizing,chen2025dynamic}, among others. Traditional approaches often rely on combinatorial optimization or rule-based heuristics. 
For instance, Zhang et al. \cite{zhang2017taxi} proposed a framework integrating combinatorial optimization with global efficiency metrics and Bayesian destination prediction derived from travel histories to improve taxi dispatch performance. However, this method requires computing all possible driver-order matches, hindering its application in large-scale scenarios due to computational complexity. 
Similarly, Seow et al. \cite{seow2009collaborative} introduced a multi-agent taxi dispatch system using collaborative negotiation among drivers, jointly optimizing passenger and driver satisfaction through simultaneous regional allocation. 
Nevertheless, these traditional approaches often struggle to scale dynamically with real-time demand fluctuations and complex urban traffic patterns, primarily due to their reliance on pre-defined heuristics or computationally intensive global matching.

While traditional methods provide foundational insights, their rigidity in dynamic environments has motivated researchers to explore data-driven RL paradigms. These approaches aim to overcome scalability constraints by learning adaptive policies through environmental interactions. 
Xu et al. \cite{xu2018large} formulated order dispatching as a bipartite graph matching problem, enabling efficient optimization of resource utilization and user experience under spatiotemporal constraints. 
Recent studies \cite{han2022real,sadeghi2022reinforcement,tang2019deep} adopt a two-stage framework combining learning and planning: first developing state value functions through training, then formulating bipartite graph models connecting drivers with potential orders during planning operations. 
Multi-agent reinforcement learning approaches \cite{zhou2021multi,li2019efficient} have demonstrated particular effectiveness in coordinating complex decision-making for large-scale order dispatching. 
In driver repositioning, Lin et al. \cite{lin2018efficient} developed a fleet management strategy that adapts to complex demand-supply dynamics, showing significant potential for real-time resource allocation. 
While prior works \cite{xu2020recommender,miao2015taxi} applied Markov Decision Processes and deep RL to driver repositioning, they fail to capture realistic supply-demand dynamics, leading to suboptimal performance. 
Zheng et al. \cite{zheng2022supply} addressed this limitation through a supply-demand-aware deep RL framework featuring a deep Q-network with action sampling mechanisms.

For joint optimization of order dispatching and driver repositioning, CoRide \cite{jin2019coride} proposed a multi-agent RL framework with hierarchical coordination, unified action space formulation, and multi-scale decision-making using attention mechanisms. 
Sun et al. \cite{sun2024optimizing} introduced JDRCL, which jointly optimizes efficiency and driver income fairness under budget constraints through group-based action representations and a primal-dual training algorithm. 
Concurrently, JDRL \cite{sun2022optimizing} employs action space grouping and an iterative minimax algorithm to balance long-term efficiency with max-min fairness, validated across 13 million real-world orders. 
Despite these advances, RL-based methods face persistent challenges in generalizing to unseen scenarios and modeling complex human behavior patterns.
LLMs offer complementary strengths through their contextual understanding, adaptability to novel situations, and interpretable reasoning capabilities, highlighting critical directions for future research.

\subsection{LLMs in Decision-Making for Transportation Systems}
LLMs have recently shown significant promise in transportation decision-making, leveraging their ability to process unstructured data \cite{liu2024spatial,gruver2024large}, reason about complex scenarios \cite{grigorev2024integrating,xue2024prompt,lai2024llmlight}, and generate human-interpretable strategies \cite{xue2023promptcast}. Early applications primarily focused on natural language interfaces for tasks such as route planning \cite{meng-etal-2024-llm,song2023llm}, traffic forecasting \cite{liu2024spatial}, and human mobility prediction \cite{xue2022leveraging}. However, recent advancements have shifted toward utilizing LLMs as core decision-making agents \cite{lai2024llmlight}.
One notable example is LLMLight \cite{lai2024llmlight}, a framework that employs LLMs for traffic signal control by integrating real-time traffic conditions through instructive prompts. Similarly, UrbanGPT \cite{li2024urbangpt} introduces a unified architecture that combines spatiotemporal dependency modeling with instruction-based fine-tuning, enhancing LLMs’ ability to capture complex feature interactions. This approach not only improves prediction accuracy under data-scarce conditions but also maintains model interpretability, a crucial factor in urban computing tasks.

The integration of LLMs with reinforcement learning has opened new frontiers in transportation optimization. 
In ride-hailing systems, Han et al. \cite{han2024gpt} introduced GARLIC (GPT-Augmented Reinforcement Learning with Intelligent Control), a novel framework for vehicle dispatching. This architecture employs multiview graphs to represent hierarchical traffic patterns while learning dynamic reward functions that incorporate individual driver behaviors. The system further integrates a GPT model trained with specialized loss functions, enabling accurate demand prediction and optimization of dispatching strategies in real-world ride-hailing operations.
Beyond predictive accuracy, the interpretability of LLM-driven decisions has become a research imperative. 
Guo et al. \cite{guo2024explainable} proposed TP-LLM, an explainable foundation model-based traffic prediction framework that integrates multimodal factors through language-based inputs, eliminating complex spatiotemporal programming while achieving state-of-the-art performance via foundation model fine-tuning.
Liang et al. \cite{liang2024exploring} propose LLM-MPE, a Large Language Model-based framework for human mobility prediction during public events, leveraging LLMs' unique capacity to process textual inputs, adapt with minimal data, and produce explainable predictions through natural language.
These studies collectively highlight how LLMs address two persistent challenges in transportation systems: improving decision-making accuracy in data-constrained environments and providing transparent explanations for model outputs. The progression from task-specific interfaces to autonomous decision agents underscores LLMs' evolving role in bridging technical complexity and operational practicality within intelligent transportation systems.

\section{Problem Formulation}
\label{Problem Formulation}

In this paper, we consider a ride-hailing platform operating within a specified urban region $\mathcal{G}$ where passengers submit ride requests and idle vehicles await dispatch. The system executes an order dispatching process at one-minute intervals to assign available vehicles to incoming passenger requests. Following the order dispatching phase, vehicles remaining idle beyond a predefined maximum idle time threshold are strategically repositioned to nearby areas to increase their likelihood of matching with future requests.

Let $T$ be the set of discrete time steps for dispatching and repositioning operations. Let $\mathcal{V}$ and $\mathcal{R}$ represent the sets of vehicles and ride requests, respectively.

\textbf{Orders:} Each ride request $r_i \in \mathcal{R}^{(t)}$ at time $t$ is characterized by its request timestamp $t_i$, pickup location $o_i$, drop-off location $d_i$, instance reward $w_i$, trip distance $td_i$, trip time $tt_i$, maximum wait time $\tau_i^{\max}$.
$x_i^r=\{t_i, o_i, d_i, td_i, tt_i, \tau_i^{\max}\}$ is the attribute feature of ride request $r_i$.

\textbf{Drivers:} Each vehicle $v_j \in \mathcal{V}^{(t)}$ at time $t$ is characterized by its current location $cl_j$, target location $tl_j$, current status (idle, or occupied) $s_j$, waited time $wt_j$, idle time $\tau_j^{\textit{idle}}$, finished order $fo_j$ and current reward $cr_j$ since completing its last order. $x_j^v=\{cl_j, tl_j, s_j, wt_j, tt_i, \tau_j^{\textit{idle}}, fo_j, cr_j\}$ is the attribute feature of driver $v_j$.

\paragraph{\textbf{Order Dispatching}}
Let $\mathcal{E}^{(t)} \subset \mathcal{R}^{(t)} \times \mathcal{V}^{(t)}$ represent the set of feasible driver-order pairs where:
\begin{equation}
(r_i, v_j) \in \mathcal{E}^{(t)} \iff \text{dist}(cl_j, o_i) \leq \tau_{dis}^{\max} \land s_j = 0
\end{equation}
where $\tau_{dis}^{\max}$ is the pickup distance threshold.
The order dispatching process can be formulated as a bipartite graph matching problem $G^{(t)} = (\mathcal{R}^{(t)}, \mathcal{V}^{(t)}, \mathcal{E}^{(t)})$.
The dispatching decision is represented by a binary assignment matrix $\mathbf{X}^{(t)} = [x_{ij}^{(t)}]$ where:
\begin{equation}
x_{ij}^{(t)} = 
\begin{cases}
1 & \text{if order $r_i$ is assigned to vehicle $v_j$ at time $t$} \\
0 & \text{otherwise}
\end{cases}
\end{equation}

Subject to the constraints:
\begin{equation}
\sum_{j: (r_i,v_j) \in \mathcal{E}^{(t)}} x_{ij}^{(t)} \leq 1, \quad \forall r_i \in \mathcal{R}^{(t)} \label{eq:order_constraint}
\end{equation}
\begin{equation}
\sum_{i: (r_i,v_j) \in \mathcal{E}^{(t)}} x_{ij}^{(t)} \leq 1, \quad \forall v_j \in \mathcal{V}^{(t)} \label{eq:vehicle_constraint}
\end{equation}
\begin{equation}
x_{ij}^{(t)} \in \{0,1\}, \quad \forall (r_i,v_j) \in \mathcal{E}^{(t)} 
\label{eq:binary_constraint}
\end{equation}

Constraint \eqref{eq:order_constraint} ensures that an order is assigned to at most one vehicle, constraint \eqref{eq:vehicle_constraint} ensures that a vehicle is assigned to at most one order, and constraint \eqref{eq:binary_constraint} enforces binary decision variables.

\paragraph{\textbf{Driver Repositioning}}
Let $\mathcal{V}_{\text{idle}}^{(t)} = \{v_j \in \mathcal{V}(t) \mid s_j = 0 \land \tau_j^{\text{idle}} > \tau^{\max}\}$ be the set of vehicles requiring repositioning.
For each $v_j \in \mathcal{V}_{\text{idle}}^{(t)}$, let $\mathcal{N}_j \subset \mathcal{G}$ be the set of candidate regions within a feasible travel distance.
The repositioning decision is represented by a matrix $\mathbf{Y}(t) = [y_{jk}(t)]$ where:
\begin{equation}
y_{jk}^{(t)} = 
\begin{cases}
1 & \text{if vehicle $v_j$ is repositioned to region $g_k \in \mathcal{N}_j$} \\
0 & \text{otherwise}
\end{cases}
\end{equation}

Subject to the constraints:
\begin{equation}
\sum_{k: g_k \in \mathcal{N}_j} y_{jk}^{(t)} = 1, \quad \forall v_j \in \mathcal{V}_{\text{idle}}^{(t)} \label{eq:repo_constraint}
\end{equation}
\begin{equation}
y_{jk}^{(t)} \in \{0,1\}, \quad \forall v_j \in \mathcal{V}_{\text{idle}}^{(t)}, \forall g_k \in \mathcal{N}_j \label{eq:repo_binary}
\end{equation}

Constraint \eqref{eq:repo_constraint} ensures that each idle vehicle is assigned to exactly one repositioning region, and constraint \eqref{eq:repo_binary} enforces binary decision variables.

\paragraph{\textbf{Joint ODDR Optimization Objective}}

The objective of the joint ODDR problem is to maximize a multi-criterion function that balances platform revenue, service quality, and operational efficiency. 
The system state at time $t+1$ is determined by the previous state and the combined effect of both decision matrices:
\begin{equation}
S(t+1) = \Gamma(S^{(t)}, \mathbf{X}^{(t)}, \mathbf{Y}^{(t)})
\end{equation}
Where $\Gamma$ represents the state transition function. 

This Joint ODDR Optimization captures two critical dependencies.
(1) Current dispatching decisions $\mathbf{X}^{(t)}$ directly influence the future distribution of idle vehicles by determining which vehicles become occupied and their subsequent locations after completing trips.
(2) Repositioning decisions $\mathbf{Y}^{(t)}$ strategically redistribute idle vehicles to areas with anticipated future demand, improving the likelihood of successful matches in subsequent time periods.

\section{Methods}
\label{Methods}

\begin{figure*}[t]
  \centering
  \includegraphics[width=1\linewidth]{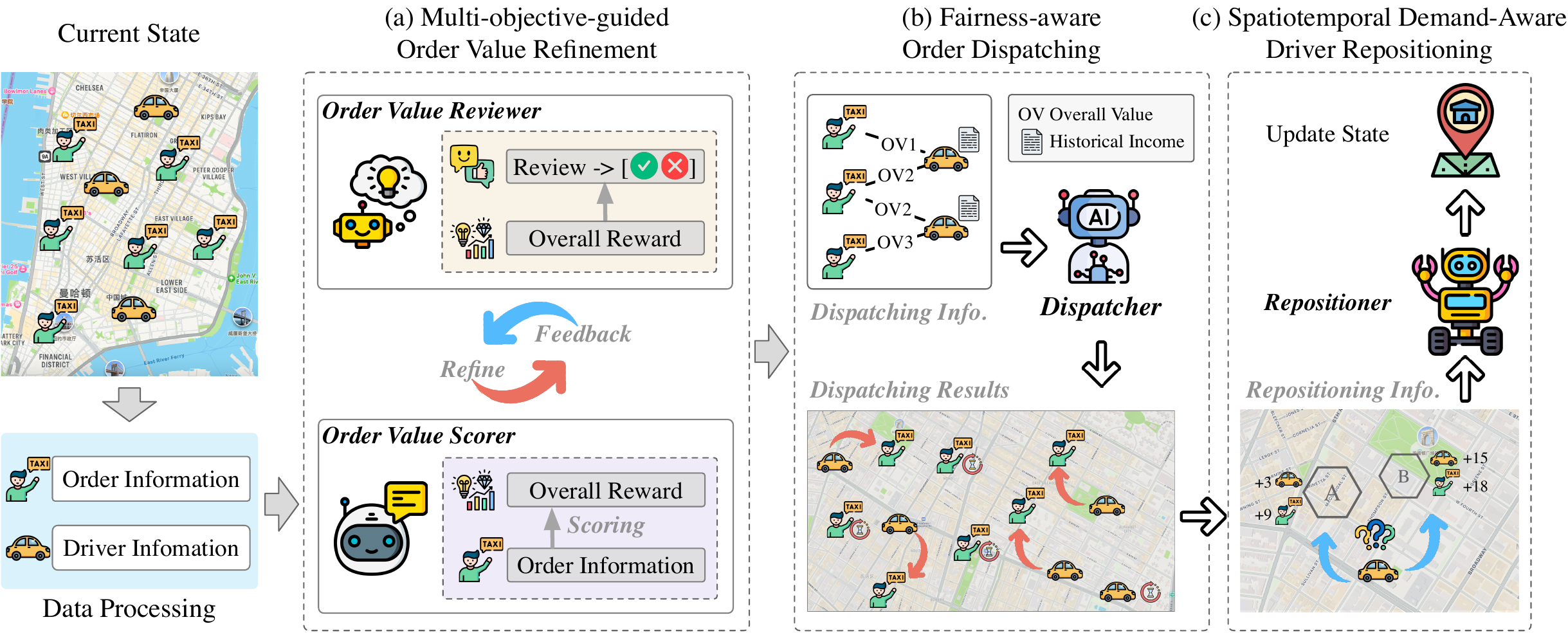}
  \caption{The proposed LLM-ODDR framework for joint order dispatching and driver repositioning tasks.}
  \label{framework}
\end{figure*}

\subsection{Overview}
The overall pipeline of the LLM-ODDR framework is illustrated in Figure \ref{framework} and Algorithm \ref{alg:llm_oddr}, comprising three major components: (1) \textit{Multi-objective-guided Order Value Refinement}, which estimates the overall value of each order using a multi-objective constraint. This component integrates an iterative self-refinement mechanism to progressively enhance scoring reasonability. (2) \textit{Fairness-aware Order Dispatching}, where orders are allocated based on the refined value scores and drivers' historical income data, ensuring fairness in the distribution process. (3) \textit{Spatiotemporal Demand-Aware Driver Repositioning}, which leverages future vehicle arrival information in a given region (derived from order dispatching outcomes) and geospatial data to optimize driver repositioning strategies. 
Additionally, we introduce \textit{Hybrid Instruction Fine-tuning}, which fine-tunes JointDR-GPT to enhance its performance in order dispatching and driver repositioning tasks.

\begin{algorithm}
\small
\caption{LLM-ODDR Framework Overview}
\begin{algorithmic}[1]
\Input Urban region $\mathcal{G}$, time set $T$, vehicles set $\mathcal{V}$, requests set $\mathcal{R}$
\Output Order dispatch decisions $\mathbf{X}$ and driver repositioning decisions $\mathbf{Y}$

\For{each time step $t \in T$}
    \State Current state: order set $\mathcal{R}^{(t)}$ and available drivers $\mathcal{V}^{(t)}$
    
    \State \textbf{Step 1: Multi-objective-guided Order Value Refinement}
    \State Initialize order values $v_i^{(0)}$ for each $r_i \in \mathcal{R}^{(t)}$
    \State Compute refined order values $v_i'$ using Algorithm \ref{alg:order_refinement}
    
    \State \textbf{Step 2: Fairness-aware Order Dispatching}
    \State Build bipartite graph $G^{(t)} = (\mathcal{R}^{(t)}, \mathcal{V}^{(t)}, \mathcal{E}^{(t)})$
    \State Compute dispatching matrix $\mathbf{X}^{(t)}$ using Algorithm \ref{alg:order_dispatch}
    \State Update $\mathcal{X}^{(t)} = \{(r_i, v_j) \mid x_{ij}^{(t)} = 1, (r_i, v_j) \in \mathcal{E}^{(t)}\}$
    
    \State \textbf{Step 3: Spatiotemporal Demand-Aware Driver Repositioning}
    \State Identify vehicles $\mathcal{V}_{\text{idle}}^{(t)} = \{v_j \in \mathcal{V}^{(t)} \mid s_j = 0 \land \tau_j^{\textit{idle}} > \tau^{\max}\}$
    \State Compute repositioning matrix $\mathbf{Y}^{(t)}$ using Algorithm \ref{alg:driver_repo}
    
    \State Update system state: $S^{(t+1)} = \Gamma(S^{(t)}, \mathbf{X}^{(t)}, \mathbf{Y}^{(t)})$
\EndFor
\end{algorithmic}
\label{alg:llm_oddr}
\end{algorithm}

Joint Optimization Mechanism: Algorithm 1 implements
a joint optimization strategy through three key mechanisms:
\begin{itemize}
    \item \text{Forward Coupling:} Dispatching decisions $\mathbf{X}^{(t)}$ directly
    determine future vehicle arrivals $V_k^{fut}$, which guides
    repositioning decisions $\mathbf{Y}^{(t)}$.
    \item \text{Backward Coupling:} Historical repositioning effectiveness 
    influences area future advantage values $f_n^{(t)} = \sum_{t'=t-T}^{t} \mathcal{N}(\mathcal{R}^{(t')}, \mathcal{D}) - \sum_{t'=t-T}^{t} \mathcal{N}(\mathcal{R}^{(t')}, \mathcal{O})$, 
    representing the net historical order density advantage of the destination area $\mathcal{D}$ over the origin area $\mathcal{O}$ within a time window $T$. This future advantage value metric in turn affects order value refinement for dispatching.
    \item \text{State Propagation:} The system state update $S^{(t+1)} = \Gamma(S^{(t)}, \mathbf{X}^{(t)}, \mathbf{Y}^{(t)})$ ensures that joint effects of both decisions propagate through time.
\end{itemize}

\subsection{Multi-objective-guided Order Value Refinement}
We begin by acquiring the current state of an experimental area $\mathcal{G}$ and processing the data to obtain order set $\mathcal{R}$ and driver information $\mathcal{V}$. To enable orders to incorporate multi-objective constraints, we developed the Multi-objective-guided Order Value Refinement module, which comprises two key components: an order value scorer and an order value reviewer.
The workflow of this multi-objective-guided refinement process is outlined in Algorithm \ref{alg:order_refinement}.

\subsubsection{Order Value Scorer}
In the order value scorer, we design a tailored prompt to instruct LLMs to evaluate the overall value of each order by considering multiple objectives. For each order $r_i \in \mathcal{R}^{(t)}$, we extract and provide order attributes $x^r_i$ including origin location $o_i$, destination location $d_i$, waiting time $t^w_i$, and immediate reward $w_i$. Additionally, we incorporate the destination area's future advantage value, which represents potential future earnings for drivers.

The future advantage value is quantified by calculating the historical order density in each region over a preceding time window:
\begin{equation}
    f^{(t)}_n = \sum_{t'=t-T}^{t}\mathcal{N}(R^{(t')}, \mathcal{D}) - \sum_{t'=t-T}^{t}\mathcal{N}(R^{(t')}, \mathcal{O})
\end{equation}
where $T$ denotes the time window (set to 60 minutes in our model), $\mathcal{D}$ and $\mathcal{O}$ denotes the destination area and original area, respectively. $R^{(t')}$ represents the set of orders at time $t'$, and $\mathcal{N}(R^{(t')}, n)$ corresponds to the number of orders at time $t'$ in region $n$.

We formulate a set of multi-objective constraints $C_{s}$ to guide the LLM's reasoning process:
\begin{equation}
    C_{s} = \{o_1, o_2, f^{(t)}_n\}
\end{equation}
where $o_1$ represents the constraint that orders with longer waiting time $t^w_i$ and higher immediate rewards $w_i$ should receive higher overall value, since prolonged waiting indicates higher service urgency and higher risk of abandonment. The constraint $o_2$ specifies that orders destined for areas with greater future advantage value $f^{(t)}_n$ should receive elevated overall value to promote long-term system efficiency and improved driver utilization.

The order value scorer then computes an overall value $v_i$ for each order by employing the LLM to reason about these objectives and constraints:
\begin{equation}
v_i = LLM(P_{scorer}, \{x^r_i, f^{(t)}_n, C_{s}\})
\end{equation}
where $P_{scorer}$ is the designed prompt template that structures the inputs and constraints for the LLM to process.

This scoring mechanism creates a holistic evaluation that balances immediate rewards with future potential, thereby addressing the platform's need to optimize for both short-term gains and long-term efficiency. The resulting order values $v_i$ are then used in subsequent order dispatching decisions.

\begin{algorithm}[t]
\small
\caption{Multi-objective-guided Order Value Refinement}
\begin{algorithmic}[1]
\Input Order set $\mathcal{R}^{(t)}$, maximum iterations $K_{max}$
\Output Refined order values $\{v_i'\}_{i=1}^{|\mathcal{R}^{(t)}|}$

\State \textbf{Initialize:} $k \gets 0$, $v_i^{(0)} \gets 0$ for all $r_i \in \mathcal{R}^{(t)}$

\While{$k < K_{max}$}
    \State \textbf{Order Value Scorer:}
    \For{each order $r_i \in \mathcal{R}^{(t)}$}
        \State Extract order attributes $x_i^r = \{t_i, o_i, d_i, td_i, tt_i, \tau_i^{\max}\}$
        \State Calculate destination region's future advantage value $f_n^{(t)}$
        \State Define multi-objective constraints:
        \State $C_{s} = \{o_1, o_2, f_n^{(t)}\}$
        \If{$k = 0$}
            \State $v_i^{(k)} \gets LLM(P_{scorer}, \{x_i^r, f_n^{(t)}, C_{s}\})$
        \Else
            \State $v_i^{(k)} \gets LLM(P_{scorer}, \{x_i^r, v_i^{(k-1)}, F_i^{(k-1)}, f_n^{(t)}, C_{s}\})$
        \EndIf
    \EndFor
    
    \State \textbf{Order Value Reviewer:}
    \State $\{s_i, F_i\}_{i=1}^{|\mathcal{R}^{(t)}|} \gets LLM(P_{reviewer}, \{x_i^r, v_i^{(k)}\}_{i=1}^{|\mathcal{R}^{(t)}|})$
    
    \If{$\sum_{i=1}^{|\mathcal{R}^{(t)}|} s_i = 0$}
        \State \textbf{break} \Comment{All valuations are reasonable}
    \EndIf
    
    \State $k \gets k + 1$
\EndWhile

\State $v_i' \gets v_i^{(k)}$ for all $r_i \in \mathcal{R}^{(t)}$
\State \Return $\{v_i'\}_{i=1}^{|\mathcal{R}^{(t)}|}$
\end{algorithmic}
\label{alg:order_refinement}
\end{algorithm}

\subsubsection{Order Value Reviewer}
To enhance the accuracy of order valuations, we designed an order value reviewer that systematically evaluates the reasonableness of the order value scorer's assessments. This module examines each order's estimated overall value and provides feedback for further refinement if necessary. The review process can be formalized as:

\begin{equation}
\{s_i, F_i\} = LLM(P_{reviewer}, \{x_i^r, v_i\}_{i=1}^{|\mathcal{R}^{(t)}|})
\end{equation}
where $P_{review}$ is the review prompt that guides the LLM to evaluate the reasonableness of all order valuations, $s_i \in \{0, 1\}$ indicates whether order $r_i$ requires reassessment (1 for reassessment, 0 for accepted), and $F_i$ represents the specific feedback for improving the valuation when $s_i = 1$. The reviewer analyzes not just individual order values in isolation, but their relative rankings and distributions across the entire set of orders.

For orders identified as requiring reassessment ($s_i = 1$), the order value scorer refines the current valuations using the provided feedback through an updated prompt. This refinement process at iteration $k+1$ can be represented as:

\begin{equation}
v_i^{(k+1)} = LLM(P_{scorer}, \{x_i^r, v_i^{(k)}, F_i^{(k)}\})
\end{equation}
where $v_i^{(k)}$ is order value at iteration $k$, and $P_{scorer}$ is the scoring prompt incorporating the reviewer's feedback $F_i^{(k)}$. This feedback guides the scorer to address specific aspects of the valuation that require adjustment, such as over-emphasizing immediate rewards or under-valuing future potential.

We implement an iterative refinement loop between the order value scorer and reviewer until one of two predefined stopping conditions is met: either (1) the maximum number of iterations $K_{max}$ is reached (set to three to avoid excessive computational costs), or (2) the reviewer deems all order valuations reasonable (formally, when $\sum_{i=1}^{|\mathcal{R}^{(t)}|} s_i = 0$). This iterative validation approach enables the LLM to progressively improve the accuracy of its evaluations through self-correction.

After meeting the stopping criteria, the final refined values $v_i^{\prime}$ represent a comprehensive balance between immediate and long-term multi-objectives, providing a solid foundation for the subsequent order dispatching phase.

\begin{algorithm}[b]
\small
\caption{Fairness-aware Order Dispatching}
\begin{algorithmic}[1]
\Input Order set $\mathcal{R}^{(t)}$ with refined values $\{v_i'\}$, available drivers $\mathcal{V}^{(t)}$
\Output Dispatching matrix $\mathbf{X}^{(t)}$

\State Initialize $x_{ij}^{(t)} \gets 0$ for all $(r_i, v_j) \in \mathcal{E}^{(t)}$

\For{each order $r_i \in \mathcal{R}^{(t)}$ sorted by $v_i'$ in descending order}
    \State Identify eligible drivers: 
    \State $\mathcal{D}_{r_i} = \{v_j \in \mathcal{V}^{(t)} \mid s_j = 0 \land \text{dist}(cl_j, o_i) \leq \tau_{dis}^{max}\}$
    
    \If{$\mathcal{D}_{r_i} \neq \emptyset$}
        \State Extract driver attributes $x_j^v$ for each $v_j \in \mathcal{D}_{r_i}$
        \State $j^* \gets LLM(P_{dispatcher}, \{r_i, v_i', \mathcal{D}_{r_i}, \{x_j^v\}_{v_j \in \mathcal{D}_{r_i}}\})$
        \State $x_{ij^*}^{(t)} \gets 1$ \Comment{Assign order $r_i$ to driver $v_{j^*}$}
        \State Update driver $v_{j^*}$ status to occupied ($s_{j^*} \gets 1$)
    \EndIf
\EndFor

\State \Return $\mathbf{X}^{(t)}$
\end{algorithmic}
\label{alg:order_dispatch}
\end{algorithm}

\subsection{Fairness-aware Order Dispatching}
Following the evaluation of each order's overall value through multi-objective-guided refinement, we conduct order dispatching while supporting the platform's sustainable development. The workflow of this part is outlined in Algorithm \ref{alg:order_dispatch}.
This dispatching process builds upon the bipartite graph matching problem $G^{(t)} = (\mathcal{R}^{(t)}, \mathcal{V}^{(t)}, \mathcal{E}^{(t)})$ defined in Section \ref{Problem Formulation}, but extends it to incorporate driver income fairness as an additional optimization objective.

For each order $r_i \in \mathcal{R}^{(t)}$, we first identify the set of eligible drivers $\mathcal{D}_{r_i} \subset \mathcal{V}^{(t)}$ based on availability and proximity:
\begin{equation}
\mathcal{D}_{r_i} = \{v_j \in \mathcal{V}^{(t)} \mid s_j = 0 \land \text{dist}(cl_j, o_i) \leq \tau_{dis}^{max}\}
\end{equation}
where $\text{dist}(cl_j, o_i)$ calculates the distance between driver location $cl_j$ and order origin $o_i$, and $\tau_{dis}^{max}$ is the maximum allowable pickup distance.

The fairness-aware order dispatching process builds upon the traditional binary assignment matrix $\mathbf{X}^{(t)} = [x_{ij}^{(t)}]$ but transforms the optimization objective to balance multiple factors including order value, pickup distance, and driver income fairness. By considering drivers' historical income distribution, the system gives preference to drivers with lower historical earnings, promoting income equality across the driver pool.

Instead of directly optimizing through traditional algorithms, we leverage the LLM's reasoning capabilities to determine the assignment:
\begin{equation}
x_{ij}^{(t)} = LLM(P_{dispatcher}, \{r_i, v^\prime_i, \mathcal{D}_{r_i}, x_j^v\}_{v_j \in \mathcal{D}_{r_i}})
\end{equation}
where $P_{dispatcher}$ is the dispatch prompt and $x_j^v$ includes driver-specific attributes such as current earnings $cr_j$, historical orders fulfilled $fo_j$, and idle time $\tau_j^{idle}$.

The resulting dispatching decisions $x_{ij}^{(t)}$ determine which orders are assigned to which drivers at time $t$, with the matched pairs denoted as:
\begin{equation}
\mathcal{X}^{(t)} = \{(r_i, v_j) \mid x_{ij}^{(t)} = 1, (r_i, v_j) \in \mathcal{E}^{(t)}\}.
\end{equation}

This formulation allows the LLM to balance immediate platform revenue (through order overall values $v^\prime_i$) with long-term driver satisfaction and retention (through consideration of historical income distribution), while enhancing decision-making interpretability by leveraging the LLM's reasoning capabilities.

\begin{algorithm}[t] 
\small
\caption{Spatiotemporal Demand-Aware Driver Repositioning}
\begin{algorithmic}[1]
\Input Idle vehicles $\mathcal{V}_{\text{idle}}^{(t)}$, dispatching results $\mathcal{X}^{(t)}$, urban region $\mathcal{G}$
\Output Repositioning matrix $\mathbf{Y}^{(t)}$
\State Initialize $y_{jk}^{(t)} \gets 0$ for all $v_j \in \mathcal{V}_{\text{idle}}^{(t)}$, $g_k \in \mathcal{G}$
\For{each vehicle $v_j \in \mathcal{V}_{\text{idle}}^{(t)}$}
    \State Identify candidate regions $\mathcal{N}_j \subset \mathcal{G}$ within two-neighborhood radius
    
    \For{each region $g_k \in \mathcal{N}_j$}
        \State Calculate historical demand: $D_k^{hist} \gets \sum_{t'=t-15}^{t} |\mathcal{R}_{g_k}^{(t')}|$
        \State Calculate historical matches: $M_k^{hist} \gets \sum_{t'=t-15}^{t} |\mathcal{X}_{g_k}^{(t')}|$
        \State Calculate future vehicle arrivals: 
        \State $V_k^{fut} \gets \sum_{(r_i,v_j) \in \mathcal{X}^{(t)}} \mathbf{1}[d_i \in g_k] \cdot f(tt_i)$
    \EndFor
    
    \State $k^* \gets LLM(P_{repo}, \{v_j, \mathcal{N}_j, \{D_k^{hist}, M_k^{hist}, V_k^{fut}\}_{g_k \in \mathcal{N}_j}\})$
    \State $y_{jk^*}^{(t)} \gets 1$ \Comment{Reposition vehicle $v_j$ to region $g_{k^*}$}
\EndFor
\State \Return $\mathbf{Y}^{(t)}$
\end{algorithmic}
\label{alg:driver_repo}
\end{algorithm}
\vspace{-0.2cm}

\subsection{Spatiotemporal Demand-Aware Driver Repositioning}
The driver repositioning module represents the culmination of our joint optimization framework, where dispatching decisions directly inform repositioning strategies through predicted vehicle arrivals, while repositioning outcomes feed back to influence future dispatching through area value updates. This bidirectional information flow ensures that both components work in concert rather than in isolation.

Following order dispatching, idle vehicles exceeding a predefined idle time threshold are repositioned to adjacent regions to enhance operational efficiency. The workflow of this idle driver repositioning is outlined in Algorithm \ref{alg:driver_repo}.
For each vehicle $v_j \in \mathcal{V}_{\text{idle}}^{(t)} = \{v_j \in \mathcal{V}^{(t)} \mid s_j = 0 \land \tau_j^{\textit{idle}} > \tau^{\max}\}$, we identify candidate repositioning regions $\mathcal{N}_j \subset \mathcal{G}$ within a two-neighborhood radius of the vehicle's current region.

To determine optimal repositioning strategies, we integrate three key spatiotemporal factors based on $\mathcal{X}^{(t)}$ that characterize the supply-demand dynamics of each candidate region $g_k \in \mathcal{N}_j$:
\begin{equation}
D_k^{hist} = \sum_{t'=t-15}^{t} |\mathcal{R}_{g_k}^{(t')}|
\label{eq:dhist}
\end{equation}
where $D_k^{hist}$ represents the total number of order requests originating from region $g_k$ in the past 15 minutes, indicating the historical demand intensity,
\begin{equation}
M_k^{hist} = \sum_{t'=t-15}^{t} |\mathcal{X}_{g_k}^{(t')}|
\end{equation}
where $M_k^{hist}$ denotes the number of successfully matched orders in region $g_k$ during the past 15 minutes, revealing the historical order fulfillment capacity.
\begin{equation}
V_k^{fut} = \sum_{(r_i,v_j) \in \mathcal{X}^{(t)}} \mathbf{1}[d_i \in g_k] \cdot f(tt_i)
\label{eq:vfut}
\end{equation}
where $V_k^{fut}$ quantifies the number of vehicles projected to arrive in region $g_k$ in the next 15 minutes based on the current dispatching decisions $\mathcal{X}^{(t)}$. Here, $\mathbf{1}[d_i \in g_k]$ is an indicator function that equals 1 if the destination of order $r_i$ belongs to region $g_k$, and $f(tt_i)$ is a time-based weighting function that accounts for the expected arrival time.

We design a specialized prompt $P_{repo}$ that encapsulates these spatiotemporal dependencies to guide the LLM in making repositioning decisions:
\begin{equation}
y_{jk}^{(t)} = LLM(P_{repo}, \{v_j, \mathcal{N}_j, \{D_k^{hist}, M_k^{hist}, V_k^{fut}\}_{g_k \in \mathcal{N}_j}\})
\end{equation}
where $y_{jk}^{(t)} \in \{0,1\}$ indicates whether vehicle $v_j$ is repositioned to region $g_k$ at time $t$. The LLM analyzes these inputs and identifies regions with high potential demand (high $D_k^{hist}$), insufficient historical fulfillment (low $M_k^{hist}$ relative to $D_k^{hist}$), and limited incoming supply (low $V_k^{fut}$) as promising candidates for repositioning.

This enables the integration of both historical patterns and future projections derived from current dispatching decisions, thereby capturing the dynamic spatiotemporal dependencies that characterize urban transportation systems. By balancing these factors, the LLM produces strategically sound repositioning decisions that optimize overall system performance while providing transparent reasoning about the decision process.

\subsection{Hybrid Instruction Fine-Tuning}
To develop a more accurate model, we utilize order dispatching and driver repositioning results (generated by GPT-4, cA2C, and TVal) to fine-tune the open-source LLM Llama3.1-70B via the Low-Rank Adaptation (LoRA) strategy \cite{hulora}, resulting in enhanced performance, named JointDR-GPT. We employ mixed-task instruction-tuning \cite{zeng2023agenttuning} to enhance the LLM's proficiency in handling both order dispatching and driver repositioning tasks.

Let $\mathcal{M}_\theta$ denote the base language model, where $\mathcal{P}_{\mathcal{M}_\theta}(y|x)$ represents the response probability distribution given instruction $x$. We consider the order dispatching task $\mathcal{T}_{dis}$ and driver repositioning task $\mathcal{T}_{rep}$. Given that both instructions and target outputs employ natural language formatting, we unify training through an end-to-end sequence-to-sequence framework. The optimization objective maximize the language model's loss:
\begin{equation}
\mathcal{L}(\theta) = \mathbb{E}_{(x,y)\sim\mathcal{T}_{dis}}[\log\mathcal{P}_{\mathcal{M}_\theta}(y|x)]+ \mathbb{E}_{(x,y)\sim\mathcal{T}_{rep}}[\log\mathcal{P}_{\mathcal{M}_\theta}(y|x)]
\end{equation}
where $x$ and $y$ represent the instruction input and instruction output of order dispatching task $\mathcal{T}_{dis}$ or driver repositioning task $\mathcal{T}_{rep}$, respectively.

\section{Experiments}
\label{Experiments}

In this section, we present a comprehensive evaluation of the LLM-ODDR framework. 
We describe the experimental settings in Section \ref{sec:Experimental Settings}, including details on datasets, baseline methods, evaluation metrics, and implementation setups. Section \ref{sec:Overall Performance} discusses the overall performance results of JointDR-GPT. Additionally, we present an ablation study in Section \ref{sec:ablation study}, analyze fairness considerations in Section \ref{sec:Fairness Study}, and provide a case study in Section \ref{sec:Case Study}.

\subsection{Experimental Settings}
\label{sec:Experimental Settings}

\subsubsection{Datasets}
\label{sec:Datasets}
The dataset utilized is the July 2015 Manhattan taxi trip records obtained from Yellow Cab's website \footnote{\url{https://www1.nyc.gov/site/tlc/about/tlc-trip-record-data.page}}. 
To evaluate the performance of JointDR-GPT across diverse scenarios, we follow the experimental settings from prior studies\cite{feng2022coordinating} by employing three datasets of varying granularities. Specifically: (1) The \textit{Large-500} dataset contains 50\% of randomly sampled orders with 500 simulated drivers; (2) The \textit{Small-200} dataset contains 10\% of randomly sampled orders with 200 simulated drivers; and (3) The \textit{Small-100} dataset comprises 10\% of randomly sampled orders paired with 100 simulated drivers. 
The simulation was conducted over five days from July 27 to July 31, 2015. This multi-granular experimental design enables robust evaluation of proposed model's performance under different scenarios.

\subsubsection{Simulator}
\label{sec:Simulator}
To evaluate our proposed JointDR-GPT and baseline methods, we utilize a ride-hailing simulator \cite{feng2022coordinating} that simulates key operational components, including order generation, order dispatching, and idle vehicle repositioning.
All experiments are conducted in a virtual environment modeled after Manhattan Island. 
The Manhattan area is partitioned into 263 similar hexagonal grids, each with a diagonal length of approximately 600 meters.
At each simulation time step $t$, the simulator provides an observation with a set of available drivers and a set of request orders. 
The time step interval $t$ is 1 minute, we assume that drivers will not move to other grids before taking a new order or receiving a repositioning command. 
At the end of each step, the simulator sequentially executes three primary tasks: (1) order overall value evaluation and review\renewcommand{\arraystretch}{0.85}
\setlength{\tabcolsep}{11pt}, (2) order dispatching, and (3) driver repositioning.
During the first stage, every newly submitted request order is evaluated for its potential overall value.
Following this, the order dispatching then generates an optimal list of driver-order pairs for the current state.
In the final stage, drivers exceeding a predefined idle time threshold are strategically relocated to enhance service efficiency. 
Once dispatching and repositioning actions are finalized, drivers actively servicing orders follow shortest-path routes to their destinations, while idle drivers remain stationary or relocate to designated grids. All matched order information is saved in a table and used to calculate the rewards for the current observation. Subsequently, the simulator returns a new observation and initiates a new cycle of order dispatching and driver repositioning.

\begin{table*}[t]
\renewcommand{\arraystretch}{1.20}
\setlength{\tabcolsep}{10pt}
\centering
\caption{Experimental results for different methods across morning, noon, and evening periods.}
\begin{tabular}{@{}ll|cc|cc|cc|cc@{}}
\toprule
\multirow{2}{*}{\textbf{Datasets}} &\multirow{2}{*}{\textbf{Method}} & \multicolumn{2}{c|}{Overall} & \multicolumn{2}{c|}{Morning (7:00-10:00)} & \multicolumn{2}{c|}{Noon (11:00-14:00)} & \multicolumn{2}{c}{Evening (17:00-20:00)}    \\ 
\cmidrule(lr){3-4} \cmidrule(lr){5-6} \cmidrule(lr){7-8} \cmidrule(lr){9-10}
& & \textbf{GMV} & \textbf{ORR} & \textbf{GMV} & \textbf{ORR} & \textbf{GMV} & \textbf{ORR} & \textbf{GMV} & \textbf{ORR} \\ 
\midrule
\multirow{10}{*}{Small-100} 
& KM                      & 50985.25 & 26.46\%  & 6550.31  & 24.16\% & 8368.05   & 26.91\% & 9581.47  & 28.34\% \\
& TVal                    & 57271.31 & 29.15\%  & 7176.96  & 22.05\% & 8686.53   & 28.79\% & 9672.29  & 30.32\% \\
& cA2C                    & 62741.94 & 31.60\%  & 9767.85  & 30.89\% & 10904.54  & 30.87\% & 9282.85  & 30.80\% \\
& GRC                     & 68056.65 & 31.65\%  & 10752.36 & 30.68\% & 10771.96  & 30.72\% & 10762.47 & 30.58\% \\
\cmidrule(lr){2-10}
& Llama3.1-8B             & 50597.33 & 24.14\%  & 6677.56  & 23.74\% & 6502.50   & 22.26\% & 6673.46  & 23.76\% \\
& Qwen2.5-72B             & 66995.29 & 29.87\%  & 10569.48 & 29.49\% & 10354.84  & 29.42\% & 10461.70 & 28.22\% \\
& Llama3.1-70B            & 68978.84 & 31.66\%  & 10679.37 & 30.51\% & 10535.12  & 30.93\% & 10180.62 & 30.44\% \\
& GPT-4o-mini             & 69076.76 & 31.99\%  & 10983.55 & 30.92\% & 10896.95  & 30.65\% & 10772.16 & 30.23\% \\
& GPT-4o                  & 69852.95 & 32.67\%  & 11339.12 & 31.77\% & 11533.46  & 31.08\% & 11460.49 & 30.17\% \\
& \textbf{JointDR-GPT}            & \textbf{70597.90} & \textbf{34.14\%}  & \textbf{11677.03} & \textbf{32.74\%} & \textbf{11902.75}  & \textbf{32.26\%} & \textbf{11673.85} & \textbf{31.76\%} \\

\midrule

\multirow{10}{*}{Small-200} 
& KM                      & 83933.19  & 50.33\%  & 13249.08 & 49.92\% & 14093.26  & 50.34\% & 14930.74 & 51.23\% \\
& TVal                    & 105432.76 & 54.32\%  & 15273.26 & 53.11\% & 15283.36  & 53.92\% & 15929.29 & 53.23\% \\
& cA2C                    & 109482.31 & 55.19\%  & 16893.46 & 54.39\% & 16921.45  & 52.10\% & 16203.36 & 55.73\% \\
& GRC                     & 111562.72 & 55.69\%  & 17856.21 & 55.37\% & 17653.14  & 56.14\% & 17954.64 & 56.74\% \\
\cmidrule(lr){2-10}
& Llama3.1-8B             & 84233.09  & 51.38\%  & 13839.84 & 48.31\% & 14048.42  & 49.20\% & 14392.44 & 50.93\% \\
& Qwen2.5-72B             & 102832.79 & 55.02\%  & 16923.49 & 54.49\% & 16932.60  & 53.94\% & 16230.85 & 53.83\% \\
& Llama3.1-70B            & 110218.62 & 55.38\%  & 17212.78 & 55.23\% & 17927.74  & 56.29\% & 17239.68 & 55.36\% \\
& GPT-4o-mini             & 112973.89 & 57.37\%  & 18028.02 & 56.20\% & 19232.19  & 58.32\% & 19223.77 & 58.44\% \\
& GPT-4o                  & 121564.51 & 58.56\%  & 18710.11 & 58.42\% & \textbf{20800.47}  & \textbf{65.02\%} & 20471.62 & 60.46\% \\
& \textbf{JointDR-GPT}            & \textbf{122074.05} & \textbf{60.03\%}  & \textbf{18990.69} & \textbf{59.27\%} & 19897.25  & 62.49\% & \textbf{21681.31} & \textbf{61.29\%} \\

\midrule

\multirow{10}{*}{Large-500} 
& KM                      & 323018.25 & 27.98\%  & 50213.96 & 28.32\% & 50231.20  & 29.13\% & 49212.70 & 27.32\% \\
& TVal                    & 342390.16 & 32.90\%  & 52392.15 & 31.53\% & 54763.06  & 34.29\% & 54322.29 & 30.32\% \\
& cA2C                    & 358003.26 & 34.91\%  & 55539.62 & 33.67\% & 55946.11  & 37.23\% & 54630.85 & 28.86\% \\
& GRC                     & 375224.15 & 35.12\%  & 54683.12 & 34.85\% & 56894.14  & 35.11\% & 57653.45 & 35.19\% \\
\cmidrule(lr){2-10}
& Llama3.1-8B             & 321326.65 & 27.14\%  & 49923.32 & 28.92\% & 49324.60  & 28.19\% & 49283.01 & 28.89\% \\
& Qwen2.5-72B             & 375210.48 & 34.02\%  & 53273.42 & 33.98\% & 54392.38  & 32.90\% & 54932.60 & 33.27\% \\
& Llama3.1-70B            & 377669.85 & 34.93\%  & 55329.63 & 34.32\% & 56832.19  & 34.93\% & 57292.91 & 34.63\% \\
& GPT-4o-mini             & 380021.96 & 35.82\%  & 58273.41 & 35.02\% & 58932.29  & 35.09\% & 59372.53 & 36.72\% \\
& GPT-4o                  & 381696.77 & 36.23\%  & 60230.19 & 35.87\% & 63492.28  & 36.23\% & 63942.94 & 36.71\% \\
& \textbf{JointDR-GPT}            & \textbf{392523.63} & \textbf{37.67\%}  & \textbf{64239.16} & \textbf{36.97\%} & \textbf{65029.71}  & \textbf{35.29\%} & \textbf{65478.27} & \textbf{37.02\%} \\ 

\bottomrule

\end{tabular}
\label{tab:results}
\end{table*}

\subsubsection{Baselines}
For traditional methods, we adopt KM as the baseline. For RL-based methods, we compare TVal\cite{xu2018large}, cA2C \cite{lin2018efficient} and GRC \cite{yang2024rethinking} with our proposed method. For LLMs, we compare with GPT-4o \footnote{\url{https://openai.com/index/hello-gpt-4o/}}, GPT-4o-mini\footnote{\url{https://openai.com/index/gpt-4o-mini-advancing-cost-efficient-intelligence/}}, the series of Llama-3.1 models\cite{dubey2024llama} and Qwen2.5 models \cite{yang2024qwen2}. 

\begin{itemize}
    \item \textbf{KM}: A greedy dispatching algorithm that constructs a bipartite graph each time slot to pair drivers and orders optimally, maximizing immediate platform efficiency.

    \item \textbf{TVal}: Integrates spatiotemporal demand-supply learning with combinatorial optimization, balancing immediate rewards and future gains for farsighted resource allocation.

    \item \textbf{cA2C}: A contextual multi-agent RL model that combines deep Q-learning and actor-critic methods, designed for efficient large-scale fleet management.

    \item \textbf{GRC}: Formulates ride-hailing dispatching as a cooperative Markov game, employing goal-reaching collaboration to optimize dispatch decisions for maximum revenue.

    \item \textbf{Qwen2.5}: Alibaba's multimodal language model with strong generative and understanding abilities, optimized across diverse tasks and domains.

    \item \textbf{Llama3.1}: Meta's latest large language model, enhanced through architectural improvements and alignment techniques, excelling in reasoning and conversational tasks.

    \item \textbf{GPT-4o-mini}: A streamlined variant of GPT-4o, optimized for performance and efficiency, suitable for cost-effective, real-time applications.

    \item \textbf{GPT-4o}: A multimodal model integrating vision and language processing, known for superior reasoning and contextual adaptability in real-world scenarios.
\end{itemize}

\subsubsection{Metrics}
Following previous studies \cite{jin2019coride,xu2018large,yang2024rethinking,sun2024optimizing}, we evaluate the performance of joint ODDR using Gross Merchandise Volume (GMV) and Order Response Rate (ORR):
\begin{itemize}
    \item \textbf{GMV:} The total value of orders fulfilled within the simulator, reflects overall transaction volume in an area.
    
    \item \textbf{ORR:} The proportion of successfully served orders to the total number of requests, which measures the platform’s operational efficiency.  
\end{itemize}

\subsubsection{Experimental Setup}
The experiments were conducted on an Ubuntu Linux server equipped with eight A800 GPUs. Following previous work \cite{feng2022coordinating}, the vehicle speed was set to 6.33 m/s, as specified in the reference. The maximum waiting time for passenger-driver matching was configured to 5 minutes, with a pick-up distance threshold of 950 meters (corresponding to a 2.5-minute travel time at the aforementioned vehicle speed). Route-driver pairs exceeding this threshold were removed from the matching process. This constraint prevents drivers from being assigned to overly distant orders, ensuring operational efficiency.

\subsection{Overall Performance}
\label{sec:Overall Performance}
We compare the performance of various models, including rule-based models, RL-based models, and models leveraging different large language models. The overall experiments performance are summarized in Table \ref{tab:results}.

As demonstrated by the experimental results, the proposed JointDR-GPT outperforms all eight baseline models across the three datasets. Specifically, regarding the overall GMV metric, the JointDR-GPT achieves improvements of 3.73\%, 12.52\%, 23.27\%, and 38.47\% compared to the GRC, cA2C, TVal, and KM models, respectively, on the small-100 dataset. On the small-200 dataset, these improvements are 9.42\%, 11.50\%, 15.78\%, and 45.44\%, respectively, while on the large-500 dataset, they are 4.61\%, 9.64\%, 14.64\%, and 21.51\%, respectively. 
These findings indicate that our proposed JointDR-GPT exhibits strong performance, effectively integrating multiple objectives into an overall scoring mechanism for each order. Consequently, it achieves more efficient order dispatching and driver repositioning.
Meanwhile, we observed that during the morning, noon, and evening peak hours, our proposed model outperforms traditional approaches. For instance, in the morning peak period on the small-100 dataset, our method improves the GMV metric by 8.60\%, 19.55\%, 62.7\%, and 78.26\% compared to the GRC, cA2C, TVal, and KM models, respectively. This demonstrates that when faced with supply-demand imbalances during peak hours, our model can adapt more quickly to the environment and make appropriate adjustments, thereby enhancing the overall operational efficiency of the platform. A detailed analysis in our case study further substantiates this finding.

Compared with other LLMs, the fine-tuned JointDR-GPT demonstrates significant improvements in overall performance. Among general-purpose models, GPT-4o achieves the highest performance, while Llama3.1-8B performs the weakest. These experimental results suggest that generalist LLMs may lack domain-specific expertise in traffic management, underscoring the need to develop specialized models for such applications. Furthermore, although LLMs employing our inference pipeline (e.g., Llama3.1-70B and Qwen2.5-72B) underperform compared to our fine-tuned JointDR-GPT, they still surpass traditional methods (KM, TVal, cA2C). These findings emphasize the benefits of multi-objective integration for holistic order evaluation in order dispatching and driver repositioning, ultimately enhancing long-term platform profitability. However, they also expose critical performance limitations, particularly in balancing regional value allocation, driver income fairness, and spatiotemporal supply-demand equilibrium. 
To bridge this gap, we propose three core innovations: Multi-Objective-Guided Order Value Refinement, Fairness-Aware Order Dispatching, and Spatiotemporal Demand-Aware Driver Repositioning. Combined with optimized JointDR-GPT, our integrated approach achieves state-of-the-art performance in joint order dispatching and driver repositioning tasks.

\begin{table}[t]
\setlength{\tabcolsep}{3.8pt}
\renewcommand{\arraystretch}{1.15}
\centering
\caption{Ablation studies of LLM-based models on two datasets.}
\begin{tabular}{@{}l|cc|cc}
\toprule
\multicolumn{1}{c|}{\multirow{2}{*}{\raisebox{-.5\height}{\centering \textbf{Models}}}} & \multicolumn{2}{c|}{Small-100} & \multicolumn{2}{c}{Large-500}    \\ 
\cmidrule(lr){2-3} \cmidrule(lr){4-5}
& \textbf{GMV} & \textbf{ORR} & \textbf{GMV} & \textbf{ORR} \\ 
\midrule
Qwen2.5-72B (w/o scorer)      & 63143.94 & 26.34\% & 370459.16 & 33.46\% \\
Qwen2.5-72B (w/o llm-dis)    & 66942.45 & 29.01\% & 372845.59 & 33.82\% \\
Qwen2.5-72B (w/o llm-rep)    & 64120.78 & 26.56\% & 370445.64 & 33.45\% \\
\textbf{Qwen2.5-72B}         & \textbf{66995.29} & \textbf{29.87\%} & \textbf{375210.48} & \textbf{34.02\%} \\
\midrule
Llama3.1-70B (w/o scorer)     & 66068.75 & 30.14\% & 372904.57 & 34.41\% \\
Llama3.1-70B (w/o llm-dis)   & 67497.59 & 31.12\% & 372660.79 & 34.26\% \\
Llama3.1-70B (w/o llm-rep)   & 66484.45 & 30.45\% & 372561.36 & 34.45\% \\
\textbf{Llama3.1-70B}        & \textbf{68978.84} & \textbf{31.66\%} & \textbf{377669.85} & \textbf{34.93\%} \\
\midrule
GPT-4o-mini (w/o scorer)           & 66246.74 & 30.31\% & 364518.32 & 34.45\% \\
GPT-4o-mini (w/o llm-dis)         & 66321.66 & 30.35\% & 364588.12 & 34.85\% \\
GPT-4o-mini (w/o llm-rep)         & 67023.15 & 31.25\% & 370136.87 & 35.10\% \\
\textbf{GPT-4o-mini}              & \textbf{69076.76} & \textbf{31.99\%} & \textbf{380021.96} & \textbf{35.82\%} \\
\midrule
GPT-4o (w/o scorer)           & 68102.69 & 30.76\% & 374921.49 & 35.61\% \\
GPT-4o (w/o llm-dis)         & 68004.86 & 30.95\% & 368451.33 & 34.65\% \\
GPT-4o (w/o llm-rep)         & 68364.97 & 31.11\% & 374648.64 & 35.44\% \\
\textbf{GPT-4o}              & \textbf{69852.95} & \textbf{32.67\%} & \textbf{381696.77} & \textbf{36.23\%} \\
\midrule
JointDR-GPT (w/o scorer)     & 68497.71 & 33.29\% & 376972.03 & 35.89\% \\
JointDR-GPT (w/o llm-dis)   & 68041.33 & 33.01\% & 369423.74 & 34.94\% \\
JointDR-GPT (w/o llm-rep)   & 68456.63 & 33.46\% & 386284.36 & 36.46\% \\
\textbf{JointDR-GPT}        & \textbf{70597.90} & \textbf{34.14\%} & \textbf{392523.63} & \textbf{37.67\%} \\
\bottomrule
\end{tabular}
\label{tab:ablation_studies}
\end{table}

\subsection{Ablation Study}
\label{sec:ablation study}
To verify the effectiveness of each component, we conduct ablation studies on three variants of different LLM models, including (1) \emph{w/o scorer} removes Multi-objective-guided Order Value Refinement strategy, only using instance rewards for order dispatching, (2) \emph{w/o llm-dis} replaces order dispatching module by the KM method, and (3) \emph{w/o llm-rep} replaces the driver repositioning module by a random choice.
The results of all variants are shown in Table~\ref{tab:ablation_studies}.

As shown in the table, JointDR-GPT and other LLM-based models consistently outperform all variants, demonstrating the efficacy of the proposed modules. Specifically, performance degrades when omitting the multi-objective-guided order value refinement strategy. This suggests that integrating multiple objectives for comprehensive order value estimation is meaningful and, in the long term, more conducive to increasing the total revenue.
We observe that \emph{w/o llm-dis} performs worse, which indicates that the dispatching module exhibits superior decision-making capabilities compared to traditional KM algorithms by comprehensively considering multiple factors, including driver fairness, thereby better facilitating sustainable platform operations. 
Furthermore, we also observe that\emph{w/o llm-rep} performs worse,  which indicates that supply-demand-based vehicle dispatch aligns more closely with operational realities and demonstrates enhanced scheduling effectiveness.
Furthermore, ablation studies reveal model-specific variations: the impact of removing \emph{llm-dis} is least pronounced in Qwen2.5-72B and Llama3.1-70B, indicating their relatively weaker inherent order dispatching capabilities. In contrast, GPT-4o-mini and GPT-4o demonstrate stronger baseline performance, while the fine-tuned JointDR-GPT achieves optimal dispatching results, confirming the effectiveness of our approach.

\begin{figure}[t]
  \centering
  \includegraphics[width=1\linewidth]{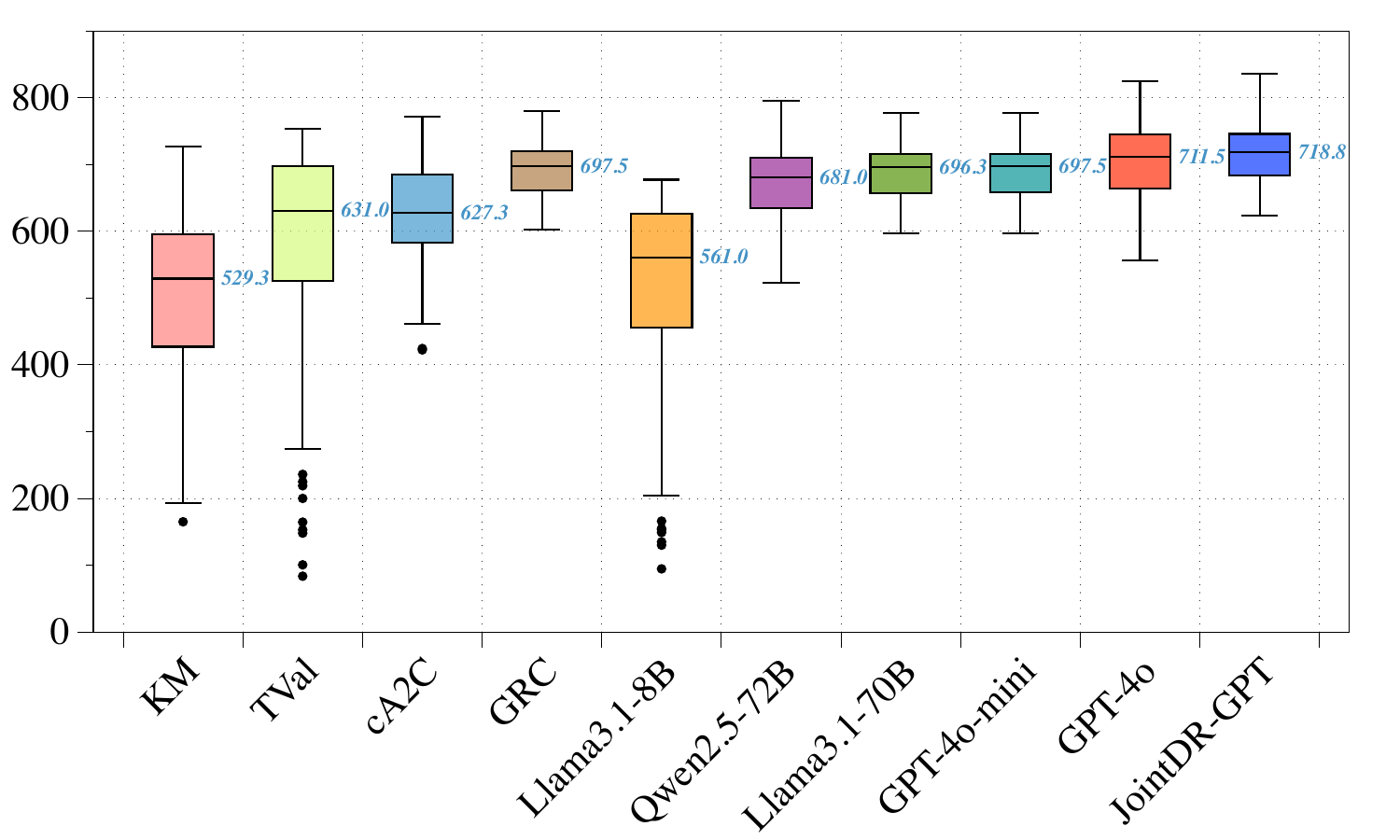}
  \caption{Fairness study of driver income for the proposed JointDR-GPT and other LLM-based models on the small-100 dataset. Annotated values indicate median income for each method.}
  \label{fig:driver_fairness}
\end{figure}

\subsection{Fairness Study}
\label{sec:Fairness Study}

We emphasize the importance of income fairness among drivers, as it fosters equitable income distribution and supports the sustainable development of platforms. To address this issue, we conducted a study on driver income fairness using the small-100 dataset, with the results presented in Fig. \ref{fig:driver_fairness}.

As shown in Fig. \ref{fig:driver_fairness}, this advantage stems from the ability of LLMs to process structured information through natural language prompts that explicitly encode fairness considerations. As shown in the dispatching prompt (Appendix A.3), driver attributes including historical income (current overall reward), number of finished orders, and idle time are presented alongside order information, with explicit instructions to consider 'ensuring fairness among drivers' as a priority factor. This prompt-based approach enables flexible integration of fairness criteria without requiring algorithmic redesign.
Additionally, the GPT-4 series achieves higher median income and GMV than the Llama3.1 series, demonstrating its superior ability to interpret and implement fairness-related instructions while enhancing dispatching efficiency. Among all LLM-based frameworks, JointDR-GPT achieved the best results, underscoring its outstanding performance in joint order dispatching and driver repositioning tasks.

\subsection{Case Study}
\label{sec:Case Study}
\begin{figure}[t]
  \centering
  \includegraphics[width=1\linewidth]{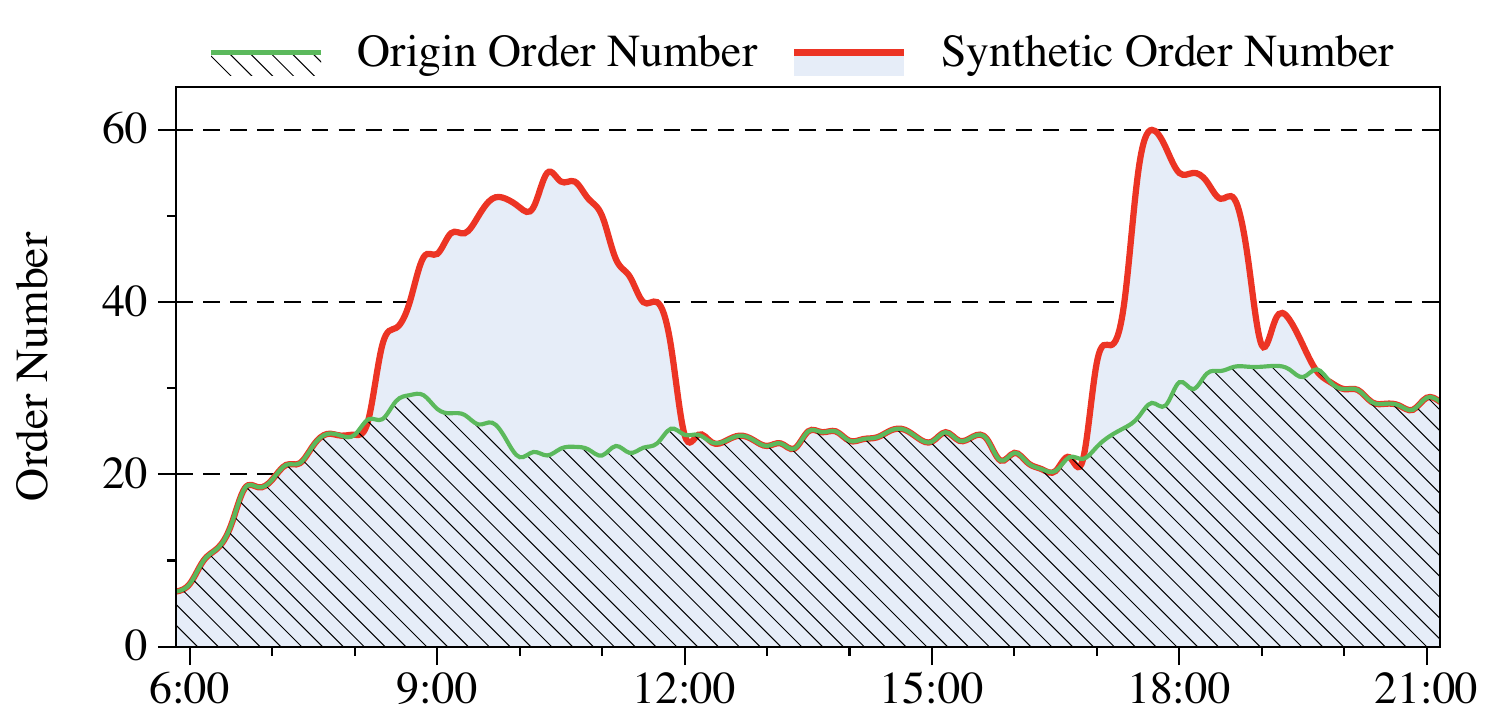}
  \caption{Comparison of order request volumes between original and synthesized datasets. Order Number represents the total number of order requests at each time point.}
  \label{fig:case_order_num}
\end{figure}

\begin{figure}[t]
  \centering
  \includegraphics[width=1\linewidth]{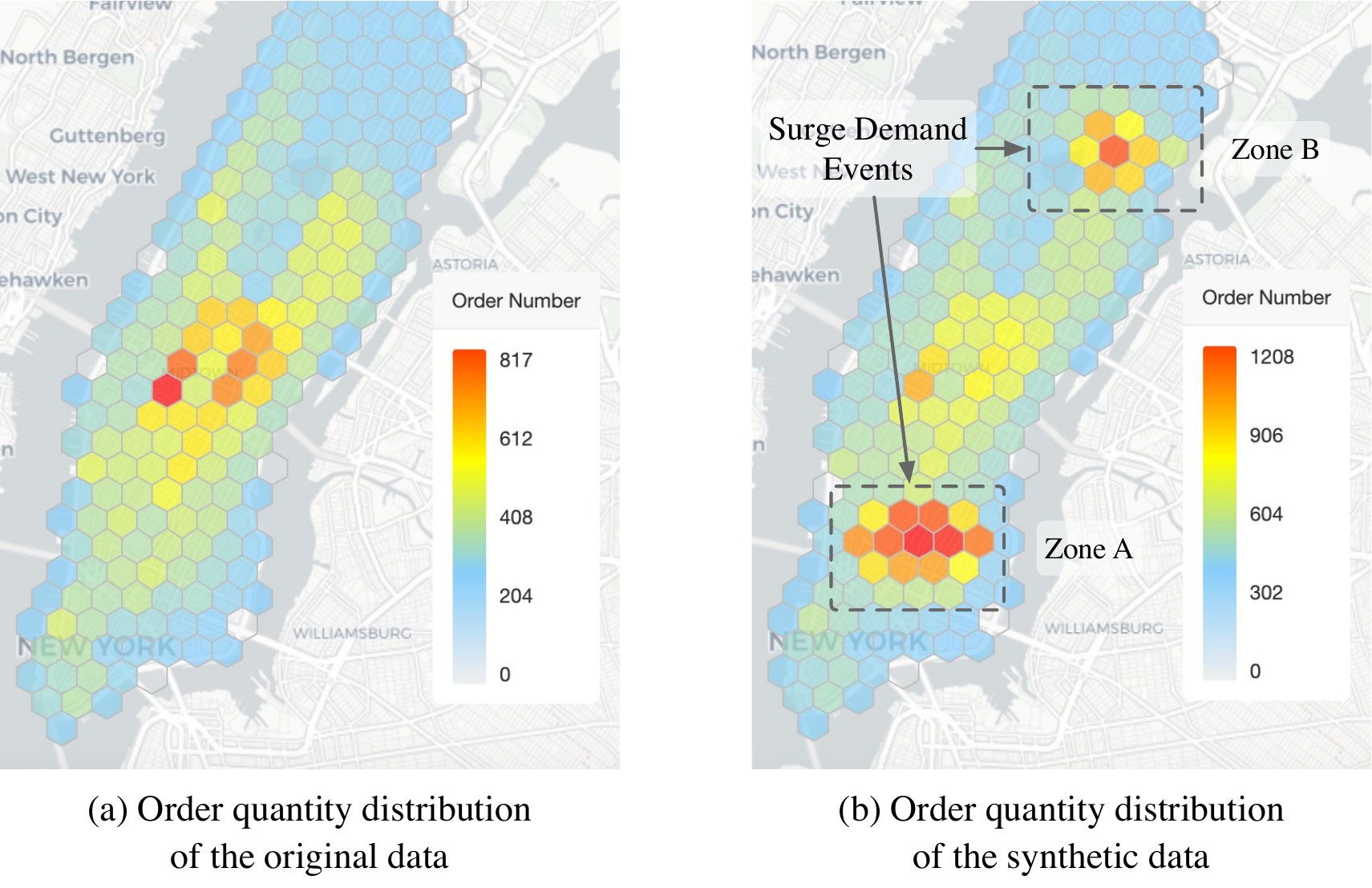}
  \vspace{-0.6cm}
  \caption{Heatmap comparison of order distribution between original dataset and synthesized dataset.}
  \label{fig:case_study_heatmap}
  \vspace{-0.4cm}
\end{figure}

To validate our framework's adaptability to surge demand events (such as concerts) and its order dispatching performance, we conducted a systematic case study. This study utilized an artificially synthesized dataset derived from the small-100 dataset, specifically designed to test model responsiveness to demand surges. For a comprehensive evaluation across different models, we allocated 300 vehicles to ensure sufficient dispatching resources.
We generated simulation data at two key time points (10:00 and 18:00) across two distinct locations. Figure \ref{fig:case_order_num} contrasts the original ride-hailing request volumes with those in the synthesized data, clearly demonstrating the dramatic increase in order demands characteristic of surge events. Figure \ref{fig:case_study_heatmap} illustrates the regional distribution comparison between original and synthesized orders, highlighting Zone A's surge event at 10:00 and Zone B's at 18:00.

\begin{figure}[t]
  \centering
  \includegraphics[width=1\linewidth]{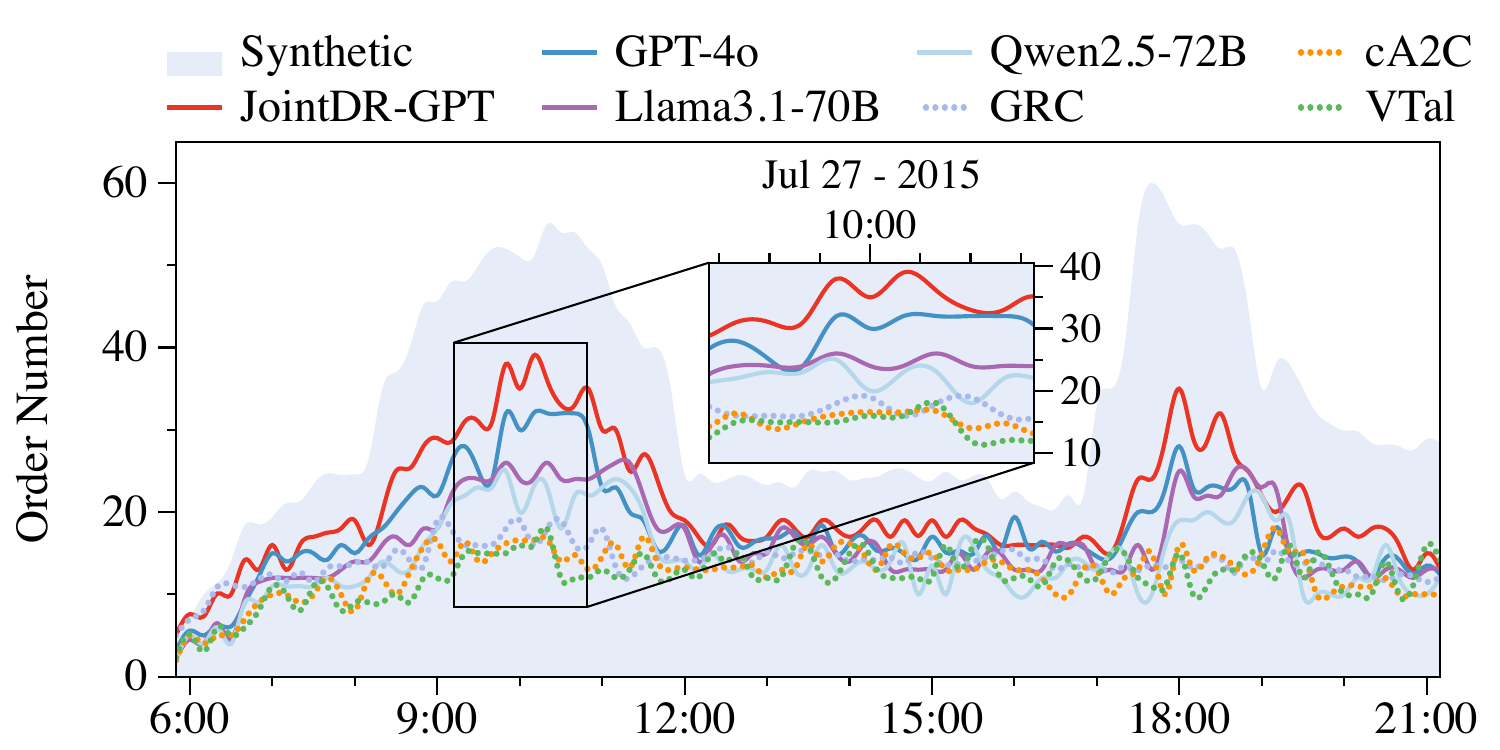}
  \caption{Performance comparison of dispatching models during surge events. Order Number denotes the total number of order requests at each time point.}
  \label{fig:case_dispatched}
\end{figure}

As shown in Figure \ref{fig:case_dispatched}, traditional models failed to effectively perceive environmental changes during these surge events, resulting in insufficient order processing capability, suboptimal matching performance, and lower order completion rates.
In contrast, our proposed large model-based methods responded positively to these demand surges through strategic adjustments. This demonstrates their ability to dynamically modify order dispatching strategies according to environmental changes. For example, upon identifying areas with higher future advantage value (based on historical order density), the system increased vehicle allocation to those locations, improving dispatching success rates. Among these advanced approaches, JointDR-GPT demonstrated superior performance, confirming its enhanced adaptability to surge demand events and more effective handling of complex dispatching tasks.

\subsection{Limitations and Discussion}
While LLM-ODDR demonstrates strong performance, we acknowledge several limitations inherent to LLM-based decision systems. First, regarding output reliability, LLMs may occasionally produce implausible decisions due to hallucination. 
Second, concerning consistency, identical inputs may occasionally yield different outputs due to the stochastic nature of LLM inference. While low temperature settings reduce this variability, complete determinism remains challenging.
Third, computational overhead is significant; LLM-based approaches require substantially more processing time than traditional methods, though batch processing considerably improves throughput for practical deployment. Importantly, the gap between LLM inference latency and the strict real-time requirements of production dispatch systems may result in increased rider cancellations and reduced platform efficiency, representing a critical barrier to deployment at scale.
Finally, our evaluation is limited to Manhattan taxi data, and generalization to other geographic regions, transportation modes, or novel scenarios not represented in training warrants further investigation. Despite these limitations, our experimental results suggest that the performance gains and interpretability benefits of LLM-ODDR justify its application in many practical scenarios, while the identified issues point to important directions for future research.

\section{Conclusion}
\label{Conclusion}
In this paper, we presented LLM-ODDR, the first LLM-based framework that jointly optimizes order dispatching and driver repositioning in ride-hailing services. By leveraging the reasoning capabilities of LLMs through our novel Multi-objective-guided Order Value Refinement, Fairness-aware Order Dispatching, and Spatiotemporal Demand-Aware Driver Repositioning components, LLM-ODDR effectively addresses the limitations of traditional approaches in handling multi-objective optimization, abnormal scenarios, and decision interpretability. Our comprehensive experiments on real-world Manhattan taxi datasets demonstrate that LLM-ODDR significantly outperforms state-of-the-art methods, including reinforcement learning approaches and other LLM-based models. The framework demonstrates exceptional adaptability to surge demand events and promotes equitable income distribution among drivers, while providing interpretable rationales for dispatching and repositioning decisions. Through the release of the pre-trained JointDR-GPT, we aim to facilitate further research in this domain and believe JointDR-GPT represents a significant advancement in leveraging LLMs for complex transportation decision-making, with potential implications for other logistics domains requiring multi-objective optimization, adaptability, and interpretability.

However, it is important to acknowledge that LLM-based approaches incur substantially higher computational costs compared to traditional methods, with inference times of approximately 17 to 30 seconds for large LLMs per decision cycle. Future work should focus on model optimization techniques such as knowledge distillation, quantization, and efficient inference strategies to reduce latency while maintaining the superior performance and interpretability benefits of LLM-based decision-making.

\section*{Acknowledgments}
This work was supported by the National Natural Science Foundation of China (Grant No. 62572417, No.92370204), the General Research Fund (GRF) of the Research Grants Council of Hong Kong, China, under Grant HKUST16207222 and PolyU15207424, the Young Scientists Fund-Type C of the National Natural Science Foundation of China (Grant No. 72501244) and CCF-DiDi GAIA Collaborative Research Funds. 

\bibliographystyle{IEEEtran}
\bibliography{document}

\vspace{-15 mm}
\begin{IEEEbiography}[{\includegraphics[width=1in,height=1.25in,clip,keepaspectratio]{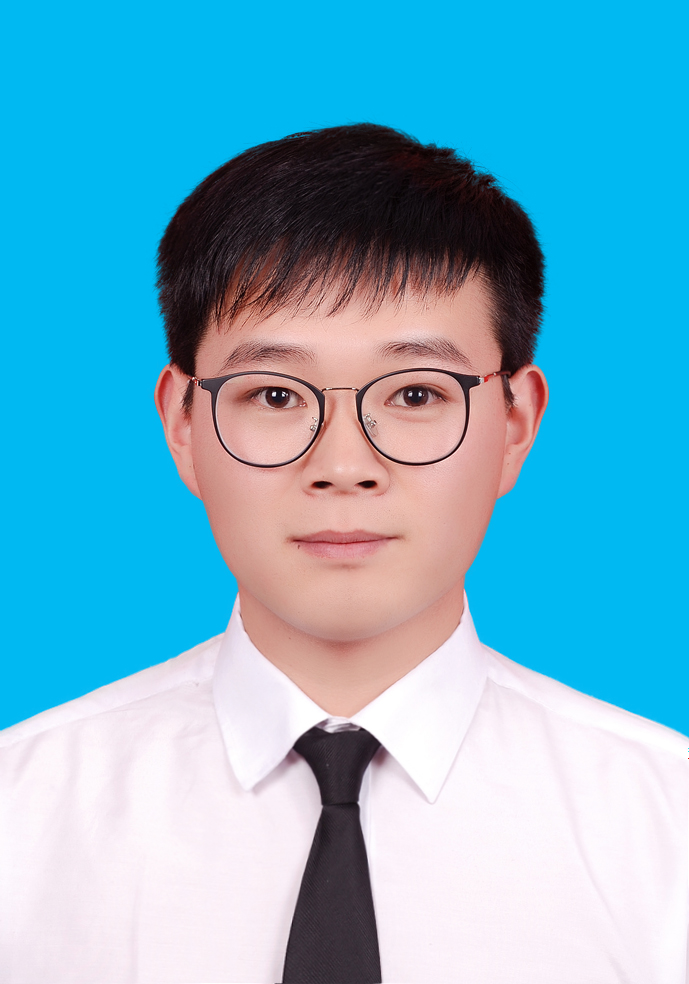}}]{Tengfei Lyu}
is a PhD student at the Thrust of Artificial Intelligence, The Hong Kong University
of Science and Technology. He received the M.S. degree from the School of Computer Science and Engineering, Central South University, Changsha, China. His research interests are AI4Science, Spatiotemporal Data Mining, and Large Language Models including their theoretical foundations and applications. He has published several research papers in prestigious conferences and journals, such
as TCBB, SIGIR, IJCAI, NeurIPS and KDD.
\end{IEEEbiography}

\vspace{-10 mm}
\begin{IEEEbiography}[{\includegraphics[width=1in,height=1.25in,clip,keepaspectratio]{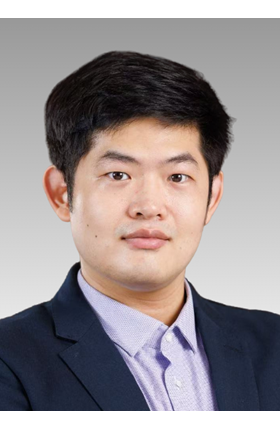}}]{Siyuan Feng}
received the B.S. degree in civil engineering from Tongji University, the M.S. degree in
civil engineering from the University of California at Berkeley, and the Ph.D. degree from the Department of Civil and Environmental Engineering, The Hong Kong University of Science and Technology. 
He is currently an Assistant Professor at the Department of Aeronautical and Aviation Engineering, and the Research Centre for Low Altitude Economy, The Hong Kong Polytechnic University, Hong Kong, China.
His research interests include economical modelling of multimodal mobility, deep learning in spatio-temporal traffic forecasting, reinforcement learning in multi-model transportation, and urban computing.
\end{IEEEbiography}

\vspace{-10 mm}
\begin{IEEEbiography}[{\includegraphics[width=1in,height=1.25in,clip,keepaspectratio]{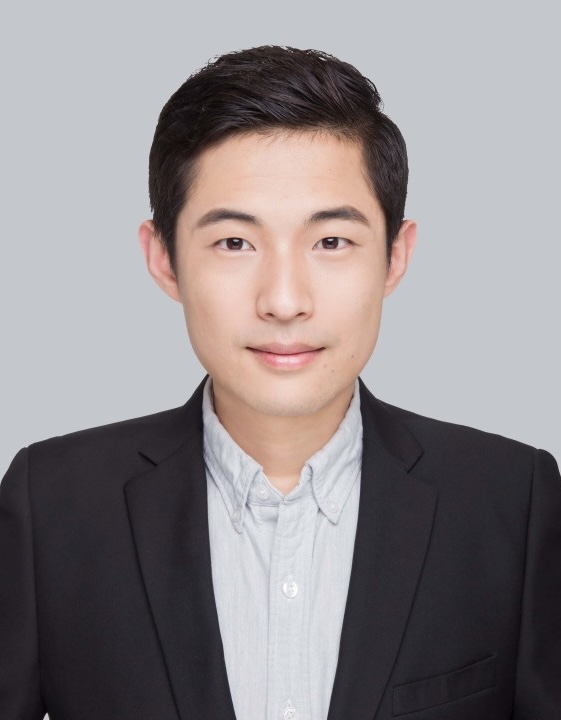}}]{Hao Liu}
received the BE degree from the South China University of Technology (SCUT), in 2012 and the PhD degree from the Hong Kong University of Science and Technology, in 2017. He is currently an Assistant Professor with the Thrust of Artificial Intelligence, The Hong Kong University
of Science and Technology (Guangzhou). Prior to that, he was a senior research scientist with Baidu Research and a postdoctoral fellow at HKUST. His general research interests are in data mining, machine learning, and Big Data management, with a special focus on mobile analytics and urban computing. He has published prolifically in refereed journals and conference proceedings, such as TKDE, KDD, ICML, NeurIPS, SIGIR, WWW, AAAI, and IJCAI.
\end{IEEEbiography}

\vspace{-10 mm}
\begin{IEEEbiography}[{\includegraphics[width=1in,height=1.25in,clip,keepaspectratio]{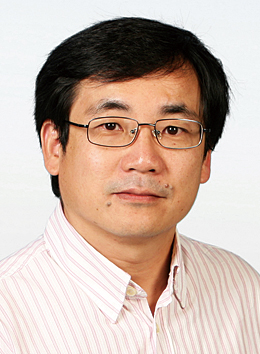}}]{Hai Yang}
is currently a Chair Professor at The Hong Kong University of Science and Technology. He is internationally known as an active scholar in the field of transportation, with more than 260 articles published in SCI/SSCI indexed journals and an H-index citation rate of 56. Most of his publications appeared in leading international journals, such as Transportation Research and Transportation Science. He is in the Distinguished Journal Editorial Board of Transportation Research Part B: Methodological, a top journal in the field of transportation.
\end{IEEEbiography}

\clearpage
\onecolumn
\appendices
\renewcommand{\thesubsection}{\Alph{section}.\arabic{subsection}}

\section{LLM Prompt Templates}
\label{appendix:prompts}

This appendix presents the complete prompt templates used in our LLM-ODDR framework. Each module employs carefully designed prompts that leverage the LLM's reasoning capabilities for ride-hailing optimization tasks. The source code is publicly available at \url{https://github.com/usail-hkust/LLM-ODDR}.

\subsection*{A.1\quad Order Value Scoring Module}
\label{appendix:scoring_prompt}

The Order Value Scoring Module evaluates each incoming order's comprehensive value by analyzing multiple factors including immediate reward, spatial characteristics, and future positioning advantage.

\vspace{0.5em}
\begin{tcolorbox}[colback=gray!5!white,colframe=gray!75!black,title=System Prompt: Order Value Scoring]
You are an expert in ride-hailing system. You can use your transportation knowledge and common sense to comprehensively assess the value of each order.
\end{tcolorbox}

\vspace{0.5em}
\begin{tcolorbox}[colback=blue!5!white,colframe=blue!75!black,title=User Prompt: Order Value Scoring,breakable]
You are an expert in ride-hailing management with extensive knowledge of transportation logistics and operational efficiency. Your task is to estimate the Value-Adding Factor (ranging between 0 and 1) for incoming orders in a ride-hailing system, balancing multiple objectives to ensure customer satisfaction and maximize overall value.

\vspace{0.8em}
\texttt{The current time: \{current\_experiment\_time\}}

\vspace{0.8em}
You will be given data on available orders in the following format:

\vspace{0.5em}
\texttt{order\_id, origin\_lng, origin\_lat, dest\_lng, dest\_lat, immediate\_reward, wait\_time, origin\_num\_1h\_ago, dest\_num\_1h\_ago, future\_value\_of\_dest\_area}

\vspace{0.3em}
\texttt{\{order\_data\}}

\vspace{0.8em}
Please analyze each order using the given data, considering the following hints:
\begin{itemize}[noitemsep,topsep=4pt,leftmargin=2em]
    \item Higher-priced orders with longer trip time and waiting times should have a higher value-adding factor.
    \item \texttt{future\_value\_of\_dest\_area = dest\_num\_1h\_ago - origin\_num\_1h\_ago}. A higher \texttt{future\_value\_of\_dest\_area} indicates that the destination has a potentially higher future advantage value, increasing the order's worth; otherwise, its value may be lower.
    \item Analyze the starting and destination coordinates along with the given timestamp to assess any additional factors that might influence the order's value (e.g., peak hours, location popularity).
\end{itemize}

\vspace{0.8em}
Please answer: For each unique order, estimate the Value-Adding Factor (between 0 and 1) that will most effectively enhance overall efficiency in subsequent order dispatching.

\vspace{0.8em}
Requirements:
\begin{itemize}[noitemsep,topsep=4pt,leftmargin=2em]
    \item Let's think step by step.
    \item Try to understand the meaning of each variable.
    \item Your estimation can only be given after finishing the analysis.
    \item Ensure that each order must estimate one Value-Adding Factor and it is between 0 and 1.
    \item For each order, your estimated Value-Adding Factor must be identified by the tag: 
    
    \texttt{$<$increase\_factor$>$ORDER\_ID:ESTIMATED\_FACTOR$<$/increase\_factor$>$}
    
    e.g., \texttt{$<$increase\_factor$>$8898691:0.273$<$/increase\_factor$>$}
\end{itemize}
\end{tcolorbox}

\subsection*{A.2\quad Self-Review Module}
\label{appendix:review_prompt}

The Self-Review Module evaluates the reasonableness of the estimated Value-Adding Factors and provides feedback for iterative refinement.

\vspace{0.5em}
\begin{tcolorbox}[colback=orange!5!white,colframe=orange!75!black,title=User Prompt: Self-Review Module,breakable]
You are an expert in ride-hailing management with extensive knowledge of transportation logistics and operational efficiency. Your task is to evaluate the estimated Value-Adding Factors (ranging between 0 and 1) provided for incoming orders in a ride-hailing system.

\vspace{0.8em}
\textbf{Your Task:} Evaluate whether the estimated Value-Adding Factors for each order are reasonable based on the last round results.

\vspace{0.8em}
Requirements:
\begin{itemize}[noitemsep,topsep=4pt,leftmargin=2em]
    \item Let's think step by step.
    \item Your evaluation can only be given after finishing the analysis.
\end{itemize}

\vspace{0.8em}
If any orders are unreasonable:
\begin{itemize}[noitemsep,topsep=4pt,leftmargin=2em]
    \item Identify and explain the reasons why the estimations are not reasonable.
    \item Provide suggestions on how to improve the estimations.
    \item Provide the identifier: \texttt{$<$evaluation\_again$>$}
\end{itemize}

\vspace{0.8em}
If all the orders are reasonable and require no further processing:
\begin{itemize}[noitemsep,topsep=4pt,leftmargin=2em]
    \item Provide the identifier: \texttt{$<$evaluation\_complete$>$}
    \item Only one identifier, either \texttt{$<$evaluation\_again$>$} or \texttt{$<$evaluation\_complete$>$}, can be provided.
\end{itemize}
\end{tcolorbox}

\subsection*{A.3\quad Order Dispatching Module}
\label{appendix:dispatching_prompt}

The Order Dispatching Module assigns orders to available drivers by considering multiple optimization objectives including reward maximization, wait time minimization, and driver fairness.

\vspace{0.5em}
\begin{tcolorbox}[colback=gray!5!white,colframe=gray!75!black,title=System Prompt: Order Dispatching]
You are an expert in ride-hailing order dispatching management. You can use your knowledge of transportation commonsense to solve this ride-hailing order dispatching tasks.
\end{tcolorbox}

\vspace{0.5em}
\begin{tcolorbox}[colback=blue!5!white,colframe=blue!75!black,title=User Prompt: Order Dispatching,breakable]
You are an expert in ride-hailing order dispatching management with extensive knowledge of transportation logistics and operational efficiency. Your task is to optimize driver assignments for incoming orders in a ride-hailing system, balancing multiple objectives to ensure customer satisfaction, operational efficiency, and driver fairness.

\vspace{0.8em}
Use your expertise to analyze the given data and make optimal dispatch decisions. When making dispatch decisions, consider the following factors in order of priority:
\begin{itemize}[noitemsep,topsep=4pt,leftmargin=2em]
    \item Maximizing immediate rewards.
    \item Minimizing customer wait times.
    \item Maximizing driver utilization.
    \item Ensuring fairness among drivers.
    \item Optimizing for long-term system efficiency.
\end{itemize}

\vspace{0.8em}
\texttt{The current time: \{current\_time\}}

\vspace{0.8em}
You will be given data on available orders and potential driver assignments in the following format:

\vspace{0.5em}
\texttt{order id: \{order\_id\}, overall immediate reward: \{immediate\_reward\_factor\}, the time already waited: \{wait\_time\}, the maximum waiting time: \{maximum\_wait\_time\}.}

\vspace{0.3em}
\texttt{Available drivers list for order \{order\_id\}:}\\
\texttt{driver id, distance from driver to order, num of finished order, current overall reward}\\
\texttt{\{driver\_list\_data\}}

\vspace{0.8em}
Please answer: For each unique order, which is the optimal driver that will most significantly improve the order dispatching tasks in the ride-hailing system?

\vspace{0.8em}
Requirements:
\begin{itemize}[noitemsep,topsep=4pt,leftmargin=2em]
    \item Let's think step by step. Your analysis should be thorough but concise.
    \item Provide your analysis for understanding the distribution and relationship between the orders and drivers.
    \item You must ensure that each driver can only be assigned to at most one order.
    \item If you determine that an order should not be assigned to any driver, explain your reasoning and disregard it.
    \item Answer your chosen driver for each order. Your choice can only be given after finishing the above analysis.
    \item For each assignment, your choice must be identified by the tag: 
    
    \texttt{$<$dispatch$>$ORDER\_ID:CHOSEN\_DRIVER\_ID$<$/dispatch$>$}
    
    e.g., \texttt{$<$dispatch$>$8898691:66$<$/dispatch$>$}
\end{itemize}
\end{tcolorbox}

\subsection*{A.4\quad Driver Repositioning Module}
\label{appendix:repositioning_prompt}

The Driver Repositioning Module guides idle drivers to high-demand areas by analyzing spatio-temporal demand patterns and supply distribution.

\vspace{0.5em}
\begin{tcolorbox}[colback=gray!5!white,colframe=gray!75!black,title=System Prompt: Driver Repositioning]
You are an expert in ride-hailing vehicle repositioning management. You can use your knowledge of transportation commonsense to solve this ride-hailing vehicle repositioning tasks.
\end{tcolorbox}

\vspace{0.5em}
\begin{tcolorbox}[colback=blue!5!white,colframe=blue!75!black,title=User Prompt: Driver Repositioning,breakable]
You are an expert in ride-hailing order dispatching management with extensive knowledge of transportation logistics and operational efficiency. Your task is to optimize vehicle repositioning strategies in a ride-hailing system, considering spatio-temporal dependencies to ensure customer satisfaction, operational efficiency, and driver fairness.

\vspace{0.8em}
\texttt{The current time: \{current\_time\}}

\vspace{0.8em}
You will be given data on current vehicle locations, potential repositioning areas, and predicted future order in the areas in the following format:

\vspace{0.5em}
\texttt{Driver: \{driver\_id\}, current longitude: \{lon\}, current latitude: \{lat\}}

\vspace{0.3em}
\texttt{Available grid for this driver:}\\
\texttt{grid id, longitude, latitude, number of order requests for this grid in the past 15 minutes, number of matched order for this grid in the past 15 minutes, number of incoming driver to this grid in the next 15 minutes}\\
\texttt{\{grid\_data\}}

\vspace{0.8em}
Please answer: For each driver in the ride-hailing system, which is the optimal grid it should be repositioned to, considering spatio-temporal dependencies, that will most significantly improve the driver repositioning tasks?

\vspace{0.8em}
Requirements:
\begin{itemize}[noitemsep,topsep=4pt,leftmargin=2em]
    \item Let's think step by step. Your analysis should be thorough but concise.
    \item You must follow the following steps to provide your analysis:
    \begin{itemize}[noitemsep,topsep=2pt,leftmargin=1.5em]
        \item Step 1: Analyze the spatio-temporal distribution and relationship between the vehicles, given grids, and traffic or regional factors over time.
        \item Step 2: Specify your chosen repositioning area for each driver.
    \end{itemize}
    \item Your choice can only be given after finishing the analysis.
    \item You must ensure that each driver is assigned to only one repositioning grid.
    \item If you determine that a vehicle should not be repositioned, explain your reasoning and disregard it.
    \item For each assignment, your choice must be identified by the tag: 
    
    \texttt{$<$reposition$>$DRIVER\_ID:CHOSEN\_GRID\_ID$<$/reposition$>$}
\end{itemize}
\end{tcolorbox}

\section{Sample Input-Output Demonstrations}
\label{appendix:samples}

This section presents concrete examples demonstrating the LLM's reasoning process for each module.

\subsection*{B.1\quad Sample: Order Value Scoring}
\label{appendix:sample_scoring}

\begin{tcolorbox}[colback=green!5!white,colframe=green!60!black,title=Sample Input: Order Value Scoring]
\small
\texttt{The current time: 2015-07-27 18:30:00}

\vspace{0.5em}
\texttt{order\_id, origin\_lng, origin\_lat, dest\_lng, dest\_lat, immediate\_reward, wait\_time, origin\_num\_1h\_ago, dest\_num\_1h\_ago, future\_value\_of\_dest\_area}

\vspace{0.3em}
\texttt{8847521, -73.9857, 40.7484, -73.9662, 40.7614, 18.50, 45, 127, 89, -38}\\
\texttt{8847522, -73.9712, 40.7831, -73.9857, 40.7484, 24.75, 30, 64, 142, 78}\\
\texttt{8847523, -73.9445, 40.7678, -73.9234, 40.8012, 31.20, 60, 45, 38, -7}\\
\texttt{8847524, -73.9923, 40.7527, -73.9556, 40.7689, 15.80, 15, 156, 67, -89}
\end{tcolorbox}

\vspace{0.5em}
\begin{tcolorbox}[colback=yellow!5!white,colframe=yellow!60!black,title=Sample Output: Order Value Scoring,breakable]
\textbf{Analysis:}

Let me analyze each order step by step, considering the key factors:

\vspace{0.5em}
\textbf{Order 8847521:}
\begin{itemize}[noitemsep,topsep=2pt,leftmargin=2em]
    \item Location: Midtown Manhattan to Upper East Side
    \item Immediate reward: \$18.50 (moderate); Wait time: 45 seconds (moderate urgency)
    \item Future value: $-$38 (negative - destination has lower demand than origin)
    \item \textit{Value-Adding Factor: Moderate-low due to negative positioning value}
\end{itemize}

\vspace{0.3em}
\textbf{Order 8847522:}
\begin{itemize}[noitemsep,topsep=2pt,leftmargin=2em]
    \item Location: Upper West Side to Midtown
    \item Immediate reward: \$24.75 (good); Wait time: 30 seconds (lower urgency)
    \item Future value: +78 (highly positive - destination is a high-demand area)
    \item \textit{Value-Adding Factor: High due to both good reward and excellent positioning}
\end{itemize}

\vspace{0.3em}
\textbf{Order 8847523:}
\begin{itemize}[noitemsep,topsep=2pt,leftmargin=2em]
    \item Location: East Harlem area
    \item Immediate reward: \$31.20 (highest); Wait time: 60 seconds (highest urgency)
    \item Future value: $-$7 (slightly negative but nearly neutral)
    \item \textit{Value-Adding Factor: High due to excellent immediate reward and urgent status}
\end{itemize}

\vspace{0.3em}
\textbf{Order 8847524:}
\begin{itemize}[noitemsep,topsep=2pt,leftmargin=2em]
    \item Location: Hudson Yards to Midtown East
    \item Immediate reward: \$15.80 (lowest); Wait time: 15 seconds (just arrived)
    \item Future value: $-$89 (highly negative)
    \item \textit{Value-Adding Factor: Low}
\end{itemize}

\vspace{0.5em}
\textbf{Estimated Value-Adding Factors:}

\texttt{$<$increase\_factor$>$8847521:0.42$<$/increase\_factor$>$}\\
\texttt{$<$increase\_factor$>$8847522:0.85$<$/increase\_factor$>$}\\
\texttt{$<$increase\_factor$>$8847523:0.78$<$/increase\_factor$>$}\\
\texttt{$<$increase\_factor$>$8847524:0.25$<$/increase\_factor$>$}
\end{tcolorbox}

\subsection*{B.2\quad Sample: Order Dispatching}
\label{appendix:sample_dispatching}

\begin{tcolorbox}[colback=green!5!white,colframe=green!60!black,title=Sample Input: Order Dispatching, breakable]
\texttt{You are an expert in ride-hailing order dispatching management with extensive knowledge of transportation logistics and operational efficiency. Your task is to optimize driver assignments for incoming orders in a ride-hailing system, balancing multiple objectives to ensure customer satisfaction, operational efficiency, and driver fairness.}\\

\texttt{Use your expertise to analyze the given data and make optimal dispatch decisions.When making dispatch decisions, consider the following factors in order of priority:}\\
\texttt{- Maximizing immediate rewards.}\\
\texttt{- Minimizing customer wait times.}\\
\texttt{- Maximizing driver utilization.}\\
\texttt{- Ensuring fairness among drivers.}\\
\texttt{- Optimizing for long-term system efficiency.}\\

\texttt{The current time: 2015-07-27 00:07:00}\\

\texttt{You will be given data on available orders and potential driver assignments in the following format.}

order id: 7943273, overall immediate reward: 3.25, the time already waited:60, the maximum waiting time: 300. \\
\texttt{Available drivers list for order 7943273:}\\
\texttt{driver id, distance from driver to order, num of finished order, current overall reward}\\
\texttt{52, 237.9924, 0, 0.0 }\\
\texttt{61, 237.9924, 0, 0.0 }\\
\texttt{69, 237.9924, 0, 0.0 }\\

order id: 7943356, overall immediate reward: 18.9, the time already waited:60, the maximum waiting time: 300. \\
\texttt{Available drivers list for order 7943356:}\\
\texttt{driver id, distance from driver to order, num of finished order, current overall reward}\\
\texttt{36, 201.4831, 1, 3.5 }\\
\texttt{79, 295.6555, 0, 0.0 }\\

order id: 7943281, overall immediate reward: 6.8, the time already waited:60, the maximum waiting time: 300. \\
\texttt{Available drivers list for order 7943281:}\\
\texttt{driver id, distance from driver to order, num of finished order, current overall reward}\\
\texttt{53, 821.0345, 0, 0.0 }\\
\texttt{90, 821.0345, 0, 0.0 }\\
\texttt{91, 821.0345, 0, 0.0 }\\
\texttt{92, 821.0345, 0, 0.0 }\\

order id: 7943287, overall immediate reward: 22.75, the time already waited:60, the maximum waiting time: 300. \\
\texttt{Available drivers list for order 7943287:}\\
\texttt{driver id, distance from driver to order, num of finished order, current overall reward}\\
\texttt{96, 809.1455, 0, 0.0 }\\

order id: 7943298, overall immediate reward: 6.75, the time already waited:60, the maximum waiting time: 300. \\
\texttt{Available drivers list for order 7943298:}\\
\texttt{driver id, distance from driver to order, num of finished order, current overall reward}\\
\texttt{36, 727.8843, 1, 3.5 }\\
\texttt{79, 726.7990, 0, 0.0 }\\

Please answer:\\
For each unique order, which is the optimal driver that will most significantly improve the order dispatching tasks in the ride-hailing system?

Requirements:\\
\texttt{- Let's think step by step. Your analysis should be thorough but concise.}\\
\texttt{- Provide your analysis for understanding the distribution and relationship between the orders and drivers.}\\
\texttt{- You must ensure that each driver can only be assigned to at most one order.}\\
\texttt{- If you determine that an order should not be assigned to any available driver, explain your reasoning and disregard it.}\\
\texttt{- Answer your chosen driver for each order. Your choice can only be given after finishing the above analysis.}\\
\texttt{- For each assignment, your choice must be identified by the tag: }\\
\texttt{$<$dispatch$>$ORDER_ID:CHOSEN_DRIVER_ID$<$/dispatch$>$, }\\
\texttt{e.g. $<$dispatch$>$8898691:66$<$/dispatch$>$.}

\end{tcolorbox}

\vspace{0.5em}
\begin{tcolorbox}[colback=yellow!5!white,colframe=yellow!60!black,title=Sample Output: Order Dispatching,breakable]
\texttt{To optimize the dispatching of drivers to orders, we will analyze each order and its available drivers based on the given priorities: maximizing immediate rewards, minimizing customer wait times, maximizing driver utilization, ensuring fairness among drivers, and optimizing for long-term system efficiency.}

\texttt{We will ensure that each driver is assigned to only one order.}

\text{Order Analysis:}

1. Order ID: 7943273\\
\texttt{Immediate Reward: 3.25; 
Available Drivers: 52, 61, 69 (all at the same distance of 237.9924);
All drivers have completed 0 orders and have a current reward of 0.0;
Since all drivers are equidistant and have no prior orders, any driver can be chosen. 
To ensure fairness, we can randomly select one of them.}

2. Order ID: 7943356\\
\texttt{
Immediate Reward: 18.9;
Available Drivers: 36 (201.4831), 79 (295.6555);
Driver 36 is closer and has completed 1 order with a current reward of 3.5, while Driver 79 has completed 0 orders;
Despite Driver 36 having completed more orders, the significant reward and closer proximity make Driver 36 the optimal choice.}

3. Order ID: 7943281\\
\texttt{
Immediate Reward: 6.8
Available Drivers: 53, 90, 91, 92 (all at the same distance of 821.0345)
All drivers have completed 0 orders and have a current reward of 0.0.
Given the large distance, assigning a driver to this order is less efficient. However, if needed, any driver can be chosen randomly for fairness.}

4. Order ID: 7943287\\
\texttt{
Immediate Reward: 22.75;
Available Driver: 96 (809.1455);
Driver 96 has completed 0 orders and has a current reward of 0.0;
Despite the large distance, the high immediate reward justifies assigning Driver 96 to this order.}

5. Order ID: 7943298\\
\texttt{
Immediate Reward: 6.75;
Available Drivers: 36 (727.8843), 79 (726.7990);
Driver 79 is slightly closer and has completed 0 orders, while Driver 36 is already assigned to another order;
Driver 79 is the optimal choice due to proximity and availability.}

\text{Driver Assignments:}

\texttt{Order 7943273: Choose any driver from 52, 61, or 69. Let's choose Driver 52 for fairness.}

\texttt{Order 7943356: Choose Driver 36.}

\texttt{Order 7943281: Due to the large distance and low reward, we may consider not assigning this order unless necessary. If needed, choose Driver 53.}

\texttt{Order 7943287: Choose Driver 96.}

\texttt{Order 7943298: Choose Driver 79.}

\text{Dispatch Decisions:}

\texttt{$<$dispatch$>$7943273:52$<$/dispatch$>$}

\texttt{$<$dispatch$>$7943356:36$<$/dispatch$>$}

\texttt{$<$dispatch$>$7943287:96$<$/dispatch$>$}

\texttt{$<$dispatch$>$7943298:79$<$/dispatch$>$}

Order 7943281 is not assigned due to inefficiency, but if required, Driver 53 can be chosen.

\end{tcolorbox}

\subsection*{B.3\quad Sample: Driver Repositioning}
\label{appendix:sample_repositioning}

\begin{tcolorbox}[colback=green!5!white,colframe=green!60!black,title=Sample Input: Driver Repositioning]
\small
\texttt{The current time: 2015-07-27 18:45:00}

\vspace{0.5em}
\texttt{Driver: 2156, current longitude: -73.9234, current latitude: 40.8012}\\
\texttt{Available grid for this driver:}\\
\texttt{grid id, longitude, latitude, requests\_15m, matched\_15m, incoming\_drivers\_15m}\\
\texttt{4521, -73.9445, 40.7678, 45, 32, 8}\\
\texttt{4522, -73.9712, 40.7831, 28, 25, 12}\\
\texttt{4523, -73.9857, 40.7484, 67, 41, 15}\\
\texttt{4524, -73.9623, 40.7567, 52, 38, 6}\\
\texttt{4525, -73.9334, 40.7912, 18, 15, 4}
\end{tcolorbox}

\vspace{0.5em}
\begin{tcolorbox}[colback=yellow!5!white,colframe=yellow!60!black,title=Sample Output: Driver Repositioning,breakable]
\textbf{Step 1: Spatio-temporal Analysis}

\textit{Computing Future Value for each grid:}
Future Value = requests\_15m $-$ matched\_15m $-$ incoming\_drivers\_15m

\begin{itemize}[noitemsep,topsep=2pt,leftmargin=2em]
    \item Grid 4521: 45 $-$ 32 $-$ 8 = \textbf{+5} (moderate opportunity)
    \item Grid 4522: 28 $-$ 25 $-$ 12 = \textbf{$-$9} (oversupply expected)
    \item Grid 4523: 67 $-$ 41 $-$ 15 = \textbf{+11} (good, but many drivers incoming)
    \item Grid 4524: 52 $-$ 38 $-$ 6 = \textbf{+8} (good opportunity, fewer competitors)
    \item Grid 4525: 18 $-$ 15 $-$ 4 = \textbf{$-$1} (low demand area)
\end{itemize}

\textit{Time Context:} 18:45 is evening rush hour - Midtown areas typically see sustained demand.

\vspace{0.5em}
\textbf{Step 2: Repositioning Decision}

Grid 4524 offers the best balance: high future value (+8) with only 6 incoming drivers.

\textbf{Recommendation: Grid 4524}

\vspace{0.5em}
\textbf{Final Repositioning Decision:}

\texttt{$<$reposition$>$2156:4524$<$/reposition$>$}
\end{tcolorbox}

\twocolumn

\section{Fine-tuning Procedures and Hyperparameters}
\label{appendix:finetuning}

This section provides a description of the fine-tuning procedures, training data construction methodology, and hyperparameter configurations employed to develop JointDR-GPT. Our approach leverages Low-Rank Adaptation (LoRA) to efficiently adapt the Llama3.1-70B base model for joint order dispatching and driver repositioning tasks.

\subsection*{C.1\quad Training Data Construction}
\label{sec:appendix_data}

The training corpus for JointDR-GPT was constructed from historical ride-hailing transaction records to ensure comprehensive coverage of diverse operational scenarios. Our data construction pipeline comprises four principal stages.

\paragraph{Stage 1: Raw Data Collection.}
We collected ride-hailing transaction records spanning a two-week period (July 13--26, 2015) from the Manhattan area of New York City. This temporal window was strategically selected to precede our evaluation period (July 27--31, 2015), thereby preventing data leakage while ensuring temporal proximity for demand pattern consistency.

\paragraph{Stage 2: Scenario Sampling and Filtering.}
Training scenarios were sampled using stratified temporal sampling to ensure coverage of all time-of-day patterns (morning peak, midday off-peak, evening peak, overnight) and all five weekday types. For each sampled time step, we retained only scenarios with at least 3 active orders and at least 3 available drivers, as trivial scenarios provide no meaningful learning signal for multi-objective reasoning. This yielded 20,160 training scenarios out of 25,200 candidate time steps (80\% retention rate). Filtering statistics are summarized in Table~\ref{tab:filter}.

\begin{table}[htbp]
\centering
\caption{Training Scenario Filtering Statistics.}
\label{tab:filter}
\begin{tabular}{lc}
\toprule
\textbf{Item} & \textbf{Value} \\
\midrule
Candidate time steps (July 13--26, 2015) & 25,200 \\
Removed: fewer than 3 active orders      & 2,847  \\
Removed: fewer than 3 available drivers  & 1,893  \\
Removed: ORR $<$ 70\% under offline Hungarian & 300 \\
Final training scenarios retained        & 20,160 \\
Retention rate                           & 80.0\% \\
\bottomrule
\end{tabular}
\end{table}

\paragraph{Stage 3: Label Generation and Quality Control.}
We apply a two-stage quality filter to ensure label reliability.

\textit{(i) Dispatching label quality.}
Only scenarios where the offline combinatorial optimization (Hungarian algorithm) achieves at least 70\% Order Response Rate are retained as dispatching supervision, ensuring labels reflect genuinely solvable instances.

\textit{(ii) Reasoning trace verification.}
GPT-4o reasoning traces are verified for structural completeness (presence of all required XML tags) and logical consistency (the stated driver assignment matches the tag-extracted decision). For GPT-4o-generated traces, any sample where parsing fails or dispatching decisions violate 1-1 matching constraints is discarded and re-queried (up to 3 retries); approximately 4.2\% of initial samples were discarded through this process. 
Traces that pass parsing but contain internally inconsistent reasoning (e.g., the analysis favors Driver A but the tag assigns Driver B) are identified by a secondary LLM-based consistency check and excluded, affecting approximately 1.8\% of samples.

\paragraph{Stage 4: Instruction Formatting and Teacher Mixing.}
Each training sample was formatted according to the prompt templates in Appendix~A, ensuring strict consistency between training and inference phases. The training corpus is constructed with the following teacher mixture (Table~\ref{tab:teacher}), determined via grid search on the validation set optimizing jointly for GMV and parsing success rate.

\begin{table}[t]
\centering
\footnotesize
\setlength{\tabcolsep}{2pt}
\caption{Teacher Mixing Proportions for Hybrid Instruction Fine-Tuning.}
\label{tab:teacher}
\begin{tabular}{llcc}
\toprule
\textbf{Teacher} & \textbf{Task} & \textbf{Proportion} & \textbf{Rationale} \\
\midrule
GPT-4o & Disp + Repo & 60\% & Rich reasoning traces for interpretability \\
KM  & Disp only            & 25\% & Provably optimal structural labels \\
cA2C         & Repo only          & 15\% & Temporally-aware repositioning signal \\
\bottomrule
\end{tabular}
\end{table}

\begin{table}[htbp]
\setlength{\tabcolsep}{4pt}
\centering
\caption{Token-Length Statistics of Training Samples.}
\label{tab:tokens}
\begin{tabular}{lcc}
\toprule
\textbf{Item} & \textbf{Input Tokens} & \textbf{Output Tokens} \\
\midrule
Mean          & 3,856 & 2,324 \\
Std.\ Dev.\   & 892   & 614   \\
Minimum       & 412   & 183   \\
Maximum       & 6,731 & 4,102 \\
Max sequence length (hard cap) & \multicolumn{2}{c}{4,096} \\
Truncated samples ($>$4,096 tokens) & \multicolumn{2}{c}{5.3\%} \\
\bottomrule
\end{tabular}
\end{table}

\begin{table*}[h]
\setlength{\tabcolsep}{6pt}
\centering
\caption{Tag Extraction Patterns and Validation Rules}
\label{tab:parsing}
\begin{tabular}{p{3.0cm}p{8.5cm}p{5.5cm}}
\toprule
\textbf{Module} & \textbf{Regex Pattern} & \textbf{Validation Constraints} \\
\midrule
Order Value Scoring & \texttt{<increase\_factor>(\textbackslash d+):([\textbackslash d.]+)</increase\_factor>} & Value $\in [0.0, 1.0]$; Order ID must exist in current batch \\
\midrule
Order Dispatching & \texttt{<dispatch>(\textbackslash d+):(\textbackslash d+)</dispatch>} & Order and Driver IDs must be valid; Driver must be eligible for order \\
\midrule
Driver Repositioning & \texttt{<reposition>(\textbackslash d+):(\textbackslash d+)</reposition>} & Driver must be idle; Grid must be within feasible travel distance \\
\midrule
Self-Review & \texttt{<evaluation\_(complete|again)>} & Exactly one evaluation tag required per response \\
\bottomrule
\end{tabular}
\end{table*}

\paragraph{Serialization of Non-Linguistic Outputs.}
RL outputs (cA2C repositioning decisions) and heuristic outputs (Hungarian algorithm matching) are serialized into natural language supervision using structured text templates.
For a repositioning decision $y_{jk}^{(t)} = 1$ (driver $j$ repositioned to grid $k$), the template takes the following form:

\begin{quote}
\textit{``Analysis: [Grid $k$ statistical description from Equations~(\ref{eq:dhist})--(\ref{eq:vfut})].
Reasoning: Grid $k$ exhibits high historical demand ($D_k^{\mathrm{hist}} = X$) and low incoming supply ($V_k^{\mathrm{fut}} = Y$), making it the optimal repositioning
target.
Decision: \texttt{<reposition>}$j$:$k$\texttt{</reposition>}''}
\end{quote}

Similarly, Hungarian algorithm dispatching assignments are serialized with explanatory text describing proximity and reward maximization. Statistical descriptions are auto-generated from spatiotemporal features, creating consistent natural language framing for purely numerical decisions.

\paragraph{Token-Length Distribution.}
Table~\ref{tab:tokens} reports the token-length statistics for training samples. The 5.3\% of samples exceeding the 4,096-token limit are handled by preserving all decision tags and truncating from the middle of the reasoning section.

\subsection*{C.2\quad Fine-tuning Methodology}

We employ Low-Rank Adaptation (LoRA) for parameter-efficient fine-tuning, which introduces trainable low-rank decomposition matrices into each layer of the transformer architecture while keeping the pre-trained weights frozen. 

The fine-tuning was conducted using the LLaMA-Factory framework on a computing cluster equipped with eight NVIDIA A800 GPUs (80GB each). Table~\ref{tab:hyperparameters} presents the complete hyperparameter configuration, which was determined through systematic grid search on the validation set.

\begin{table}[h]
\centering
\caption{Fine-tuning Hyperparameters}
\label{tab:hyperparameters}
\begin{tabular}{ll}
\toprule
\textbf{Hyperparameter} & \textbf{Value} \\
\midrule
\multicolumn{2}{l}{\textit{Model Configuration}} \\
Base model & Llama3.1-70B \\
Fine-tuning method & LoRA \\
LoRA rank ($r$) & 16 \\
LoRA alpha ($\alpha$) & 32 \\
LoRA dropout & 0.05 \\
Target modules & q\_proj, k\_proj, v\_proj, o\_proj \\
\midrule
\multicolumn{2}{l}{\textit{Optimization}} \\
Optimizer & AdamW \\
Learning rate & $1 \times 10^{-4}$ \\
LR scheduler & Cosine annealing \\
Warmup ratio & 0.1 \\
Weight decay & 0.01 \\
Gradient clipping & 1.0 \\
\midrule
\multicolumn{2}{l}{\textit{Training Configuration}} \\
Batch size (per GPU) & 4 \\
Gradient accumulation steps & 4 \\
Effective batch size & 128 \\
Training epochs & 3 \\
Max sequence length & 4,096 tokens \\
Mixed precision & BF16 \\
\bottomrule
\end{tabular}
\end{table}

\textbf{Hyperparameter.} The learning rate of $1 \times 10^{-4}$ was selected as the optimal trade-off between convergence speed and training stability. Lower rates ($5 \times 10^{-5}$) resulted in slower convergence without accuracy improvements, while higher rates ($5 \times 10^{-4}$) led to training instability. The LoRA rank of 16 provides sufficient expressiveness for our domain adaptation task; experiments with rank 8 showed 2.3\% lower accuracy, while rank 32 provided negligible improvements at doubled memory cost.

\section{Output Parsing and Tag Extraction}
\label{appendix:parsing}

To integrate LLM outputs into the operational ride-hailing system, we employ a robust structured parsing mechanism that extracts actionable decisions from natural language responses. This section details the parsing strategy, validation procedures, and failure handling mechanisms.

\subsection*{D.1\quad Structured Output Format Design}

Each LLM module produces outputs containing both natural language reasoning and structured decision tags. We adopt an XML-style tagging convention that enables reliable extraction while preserving the interpretability benefits of natural language explanations. The general output structure follows this pattern:

\begin{verbatim}
[Step-by-step analysis and reasoning...]
Final Decisions:
<dispatch>ORDER_ID:DRIVER_ID</dispatch>
...
\end{verbatim}

This hybrid design offers several advantages over purely structured (JSON) or purely unstructured outputs.

\textbf{Separation of Concerns}: The reasoning component provides transparency and auditability, while decision tags enable unambiguous action extraction.
\textbf{Robust Parsing}: XML-style tags with explicit delimiters are less susceptible to parsing errors than JSON, which can fail due to unescaped characters or formatting inconsistencies common in LLM outputs.
\textbf{Partial Result Recovery}: Unlike JSON parsing which fails entirely on malformed input, our tag-based approach can extract valid decisions even when some outputs are malformed.
\textbf{Graceful Degradation}: When parsing fails completely, the system falls back to baseline methods (KM algorithm) to ensure operational continuity.

\begin{table}[h]
\setlength{\tabcolsep}{6pt}
\centering
\caption{Parsing Success Rates (\%)}
\label{tab:parsing_stats}
\begin{tabular}{lccc}
\toprule
\textbf{Model} & \textbf{Dispatching} & \textbf{Repositioning} & \textbf{Value Scoring} \\
\midrule
Llama3.1-8B & 94.2 & 91.8 & 96.1 \\
Llama3.1-70B & 98.7 & 97.9 & 99.2 \\
GPT-4o-mini & 99.1 & 98.4 & 99.5 \\
GPT-4o & 99.6 & 99.2 & 99.8 \\
JointDR-GPT & 99.8 & 99.6 & 99.9 \\
\bottomrule
\end{tabular}
\end{table}

\subsection*{D.2\quad Regular Expression Patterns and Validation}

Table~\ref{tab:parsing} presents the regular expression patterns employed for each module. All patterns incorporate validation constraints to reject malformed or out-of-range values.

\subsection*{D.3\quad Parsing Reliability and Error Handling}

We evaluated parsing reliability across our experimental runs. Table~\ref{tab:parsing_stats} reports the parsing success rates for each model.
JointDR-GPT achieves near-perfect parsing success rates ($>$99.5\%) due to explicit format training during fine-tuning. For the rare cases where parsing fails, we implement a three-tier fallback strategy.
Retry with Explicit Format Reminder: Append a format reminder to the prompt and retry inference.
Partial Extraction: Extract successfully parsed decisions and apply baseline methods only to unparsed entities.
Full Fallback: Revert to KM algorithm for the entire batch if critical failures persist.

In our experiments, Tier 1 recovery succeeded in 87\% of initial parsing failures, Tier 2 handled 11\%, and only 2\% required full fallback to baseline methods.

\section{Inference Latency and Deployment Cost}
\label{sec:appendix_latency}

To provide quantitative context for the computational overhead of LLM-based approaches, we conducted a systematic latency measurement on our evaluation server (8~$\times$ NVIDIA A800 80\,GB GPUs) using the Small-100 dataset as a representative workload. Each decision cycle covers one 1-minute simulation step including order value scoring, order dispatching, and driver repositioning. Results are averaged over 300 decision cycles across five simulation days (July 27--31, 2015) and reported in Table~\ref{tab:latency}.

\begin{table}[htbp]
\setlength{\tabcolsep}{5pt}
\centering
\caption{Inference Latency and Throughput per Decision Cycle.}
\label{tab:latency}
\begin{tabular}{lcc}
\toprule
\textbf{Method} & \textbf{Avg.\ Latency/Cycle} & \textbf{Throughput (cycles/min)} \\
\midrule
KM (baseline)      & 0.1\,s  & $\sim$600  \\
TVal               & 1\,s    & $\sim$60   \\
cA2C / GRC         & 1\,s    & $\sim$60   \\
Llama3.1-8B        & 4.2\,s  & $\sim$14   \\
Llama3.1-70B       & 18.6\,s & $\sim$3.2  \\
GPT-4o-mini (API)  & 26.1\,s & $\sim$2.3  \\
GPT-4o (API)       & 30.3\,s & $\sim$2.0  \\
JointDR-GPT (ours) & 16.8\,s & $\sim$3.6  \\
\bottomrule
\end{tabular}
\end{table}

As shown in Table~\ref{tab:latency}, LLM-based approaches incur substantially higher latency than classical optimizers and RL-based methods. Among all LLM-based methods, our fine-tuned JointDR-GPT achieves the lowest latency at 16.8\,seconds per cycle, which is 44.6\% faster than GPT-4o (30.3\,s) and 35.6\% faster than GPT-4o-mini (26.1\,s). This efficiency gain is attributable to the LoRA fine-tuning process, which specializes the model for ODDR tasks and produces more concise, structured outputs with fewer redundant reasoning tokens compared to general-purpose API models. Meanwhile, the smaller Llama3.1-8B achieves only 4.2\,s latency but, as shown in Table~\ref{tab:results}, its decision quality is substantially inferior, underscoring that raw inference speed alone is insufficient without adequate reasoning capacity. Traditional methods (KM at 0.1\,s, TVal and cA2C/GRC at approximately 1\,s) remain significantly faster, reflecting the fundamental compute gap between neural language model inference and classical optimization.

Despite the higher absolute latency, we argue that the 16.8-second overhead of JointDR-GPT is operationally acceptable for the following reasons. First, ride-hailing platforms execute dispatching at 1-minute intervals; the 16.8-second latency of JointDR-GPT consumes only 28\% of the decision window, leaving ample time for system state updates and driver communication. Second, with batch-parallel inference where order scoring, dispatching, and repositioning are pipelined across CUDA streams, the wall-clock overhead reduces to approximately 8 to 12\,seconds per step. Third, the future advantage value $f_n^{(t)}$, historical demand statistics $D_k^{\text{hist}}$, and projected vehicle arrivals $V_k^{\text{fut}}$ are computed asynchronously in the background, further decoupling LLM inference from time-critical operations. Fourth, JointDR-GPT generates high-quality reasoning traces that can serve as teacher signals for distilling a compact specialist model (e.g., 7B parameters with INT4 quantization), targeting 2 to 3\,s latency while retaining over 95\% of the performance gains.

\paragraph{Intra-batch Parallel Bipartite Decomposition.}
A practically compelling path to sub-4-second latency is to exploit the spatial structure of the dispatching problem. The full bipartite graph $\mathcal{G}^{(t)} = (\mathcal{R}^{(t)}, \mathcal{V}^{(t)}, \mathcal{E}^{(t)})$ can be partitioned into $K$ spatially disjoint sub-graphs $\mathcal{G}_1, \ldots, \mathcal{G}_K$ by clustering orders and drivers into geographic zones.
Each sub-graph is dispatched to an independent LLM inference worker in parallel; because the sub-graphs are non-overlapping by construction, the 1-1 matching constraint is automatically preserved without post-processing.
The effective per-cycle latency is thereby reduced from $L$ to approximately $L/K$.

To validate this strategy empirically, we partitioned Manhattan into $K \in \{1, 2, 4, 8, 16\}$ geographic zones and measured effective end-to-end latency, partition loss rate (cross-zone feasible pairs discarded), and GMV relative to the single-partition baseline. Results are reported in Table~\ref{tab:partition}.

\begin{table}[htbp]
\footnotesize
\setlength{\tabcolsep}{3pt}
\centering
\caption{Effect of Intra-batch Parallel Bipartite Decomposition on
JointDR-GPT (Small-100 dataset). $\star$ denotes the recommended
production configuration.}
\label{tab:partition}
\begin{tabular}{ccccc}
\toprule
\textbf{Zones ($K$)} & \textbf{Latency/Cycle} & \textbf{Speedup} & \textbf{Partition Loss} & \textbf{GMV (relative)} \\
\midrule
1 (baseline) & 16.8\,s        & $1.0\times$  & 0.0\% & 100.0\% \\
2            & 9.1\,s         & $1.85\times$ & 1.2\% & 99.4\%  \\
4            & 4.8\,s         & $3.5\times$  & 3.1\% & 98.3\%  \\
8$\star$     & \textbf{2.6\,s}& \textbf{$6.5\times$} & \textbf{4.7\%} & \textbf{97.8\%} \\
16           & 1.4\,s         & $12.0\times$ & 5.3\% & 95.8\%  \\
\bottomrule
\end{tabular}
\end{table}

At $K{=}8$ zones, effective latency drops to \textbf{2.6\,seconds}, well within the 4-second production dispatch window, while retaining 97.8\% of the single-partition GMV. Cross-zone pairs are handled by a lightweight KM boundary-reconciliation step adding less than 0.1\,s of overhead. At $K{=}16$, latency falls below 2\,s but partition loss grows more steeply (5.3\%), causing a more noticeable GMV degradation ($-$4.2\% relative). We therefore recommend $K{=}8$ as the practical sweet spot.

\paragraph{Estimated Deployment Cost.}
For API-based deployment with GPT-4o, the estimated cost is approximately \$0.04 to \$0.07\,USD per decision cycle based on current OpenAI pricing ($\sim$3,800 input tokens and $\sim$1,200 output tokens per cycle). Over a full simulated day ($\sim$1,440 cycles), this amounts to roughly \$60 to \$80\,USD, which is negligible relative to the revenue uplift: on the Large-500 dataset, JointDR-GPT increases GMV by \$16,826 over GRC (from \$375,224 to \$392,524), yielding a revenue-to-cost ratio exceeding 200:1. For self-hosted deployment, JointDR-GPT running on 8~$\times$ A800 GPUs incurs a one-time infrastructure cost, with per-cycle inference cost effectively limited to electricity consumption. Notably, the self-hosted JointDR-GPT not only achieves the best task performance (Table~\ref{tab:results}) but also operates at 16.8\,s per cycle, nearly half the latency of API-based GPT-4o (30.3\,s), making it the most cost-effective and latency-efficient option among all high-performing LLM configurations.

\section{Fairness Analysis}
\label{sec:appendix_fairness}
We emphasize the importance of income fairness among drivers, as it fosters equitable income distribution and supports the sustainable development of platforms.

\paragraph{Fairness Metric Definition.}
We formally adopt the \textit{minimum income} (max-min fairness criterion) as our primary fairness metric, consistent with the ride-hailing fairness literature~\cite{sun2022optimizing, sun2024optimizing}. Given $N$ drivers with cumulative income $I_j$ over the evaluation period, the Fairness Score is defined as:
\begin{equation}
    \mathrm{FS} = \min_{j=1,\ldots,N} I_j.
    \label{eq:fairness}
\end{equation}
A higher FS indicates that even the lowest-earning driver achieves a better income floor. We additionally report the \textit{median income} and \textit{mean income} as complementary central tendency metrics, and the \textit{income dispersion ratio} (IDR\,$=$\,$(I_{\max} - I_{\min}) / \bar{I}$) as a normalized measure of income spread, where lower IDR values indicate more equitable distribution.

\paragraph{Results.}
Table~\ref{tab:fairness_full} reports FS, median income, mean income, IDR, and GMV for all methods on the Small-100 dataset. Figure~\ref{fig:driver_fairness} visualizes the income distribution via box plots, with annotated values indicating the median income for each method.

\begin{table}[t]
\setlength{\tabcolsep}{8pt}
\centering
\caption{Fairness and Efficiency Metrics on the Small-100 Dataset.
FS: minimum driver income (Eq.~\ref{eq:fairness}); IDR: income dispersion ratio (lower is more equitable).}
\label{tab:fairness_full}
\begin{tabular}{lccccc}
\toprule
\textbf{Method} & \textbf{Min} & \textbf{Median} & \textbf{Mean} & \textbf{IDR} & \textbf{GMV} \\
\midrule
KM              & 165.0   & 529.25  & 509.85  & 1.101 & 50,985  \\
TVal            & 83.5    & 631.00  & 575.62  & 1.164 & 57,271  \\
cA2C            & 422.5   & 627.25  & 627.41  & 0.557 & 62,742  \\
GRC             & 503.0   & 697.50  & 691.88  & 0.400 & 68,057  \\
\midrule
Llama3.1-8B     & 94.5    & 561.00  & 511.75  & 1.139 & 50,597  \\
Qwen2.5-72B     & 523.0   & 681.00  & 669.95  & 0.407 & 66,995  \\
Llama3.1-70B    & 596.5   & 696.25  & 689.78  & 0.262 & 68,979  \\
GPT-4o-mini     & 596.5   & 697.50  & 690.76  & 0.262 & 69,077  \\
GPT-4o          & 557.0   & 711.50  & 702.51  & 0.382 & 69,853  \\
JointDR-GPT     & \textbf{623.5} & \textbf{718.75} & \textbf{716.03} & \textbf{0.297} & \textbf{70,598}  \\
\bottomrule
\end{tabular}
\end{table}

\paragraph{Analysis.}
The results in Table~\ref{tab:fairness_full} reveal several important findings.

First, JointDR-GPT achieves the best performance across all three income level metrics (FS\,$=$\,623.5, median\,$=$\,718.75, mean\,$=$\,716.03) while simultaneously achieving the highest GMV (70,598). This demonstrates that efficiency and fairness are jointly achievable through multi-objective LLM reasoning, rather than being fundamentally at odds.

Second, the income dispersion analysis reveals a clear separation between methods that effectively encode fairness and those that do not. Methods without explicit fairness mechanisms exhibit severe income inequality: TVal produces the highest IDR (1.164) with a minimum income of only 83.5, meaning its lowest-earning driver receives less than 15\% of the median income. KM (IDR\,$=$\,1.101, min\,$=$\,165.0) and Llama3.1-8B (IDR\,$=$\,1.139, min\,$=$\,94.5) show similarly extreme disparities, indicating that these approaches leave some drivers severely underutilized while concentrating orders on others. In contrast, all LLM-based methods with 70B+ parameters achieve IDR values below 0.41, demonstrating that adequate model capacity is essential for encoding fairness reasoning from prompt-level instructions.

Third, among the high-performing methods, Llama3.1-70B and GPT-4o-mini achieve the lowest IDR (both 0.262), reflecting the tightest income range. JointDR-GPT has a moderately higher IDR (0.297), which reflects its dispatching strategy: by proactively routing drivers to high-value areas through spatiotemporal repositioning, JointDR-GPT raises the income ceiling (max\,$=$\,836) beyond that of Llama3.1-70B and GPT-4o-mini (max\,$=$\,777.5), creating more opportunities for top-performing drivers. Crucially, it simultaneously raises the income floor to 623.5, which is the highest among all methods and 24.0\% above GRC (503.0). This means that even the lowest-earning driver under JointDR-GPT earns more than the lowest-earning driver under any other method. From the max-min fairness perspective, JointDR-GPT therefore achieves a strict Pareto improvement: no driver is worse off in the worst case, and the overall system generates significantly more revenue.

Fourth, it is noteworthy that GPT-4o, despite achieving the second-highest GMV (69,853), has a relatively higher IDR (0.382) and a lower FS (557.0) compared to Llama3.1-70B and GPT-4o-mini. This suggests that stronger general-purpose reasoning does not automatically translate to better fairness outcomes; rather, the balance between efficiency-seeking and fairness-encoding depends on the specific model's response to prompt instructions. JointDR-GPT, through fine-tuning on ODDR-specific data that explicitly includes fairness-aware decisions, achieves the most favorable combination of both objectives.

As shown in the dispatching prompt (Appendix~A.3), driver attributes including historical income (current overall reward), number of finished orders, and idle time are presented alongside order information, with explicit instructions to consider ``ensuring fairness among drivers'' as a priority factor. This prompt-based approach enables flexible integration of fairness criteria without requiring algorithmic redesign. Among all frameworks evaluated, JointDR-GPT achieves the best results on efficiency (GMV), fairness floor (FS), and income level (median and mean), underscoring its outstanding performance in joint order dispatching and driver repositioning tasks.

\end{document}